\documentclass[journal]{IEEEtran} % regular single-column

\newcommand{\final}{1}

\usepackage{color}
\usepackage[dvipsnames]{xcolor}
\usepackage{url}
\usepackage{amsfonts}
\usepackage{amssymb}
\usepackage{amstext}
\usepackage{amsmath}
\usepackage{mathtools}
\usepackage{xspace}
\usepackage{xparse}
\usepackage{theorem}
\usepackage{thmtools}
\usepackage{epsfig}
\usepackage{overpic}
\usepackage{float}
\usepackage{empheq}
\usepackage{booktabs,tabularx}
\usepackage{subcaption}
\usepackage{multirow}
\usepackage[hidelinks]{hyperref}
\usepackage[numbers]{natbib}
\hypersetup{
    colorlinks,
    linkcolor={Black},
    citecolor={Sepia},
    urlcolor={BlueViolet}
}
\usepackage[nameinlink,noabbrev]{cleveref}
\usepackage{algorithmic}
\usepackage[ruled]{algorithm2e}
\usepackage{tikz}
\usetikzlibrary{positioning,fit,calc}
\tikzset{block/.style={draw,thick,text width=4cm,minimum height=1.5cm,align=center},
     line/.style={-latex,thick}
}
\definecolor{figred}{rgb}{1,0,0}
\definecolor{figgreen}{rgb}{0,0.6,0}
\definecolor{figblue}{rgb}{0,0,1}
\definecolor{figpink}{rgb}{1,0.63,0.63}

\newcommand*\widefbox[1]{\fbox{\hspace{2em}#1\hspace{2em}}}

\NewDocumentCommand\FloatBox{ s O{.8\linewidth} m }{%
    \IfBooleanTF{#1}{\begin{figure*}[h!t]}{\begin{figure}[h{!t}]}%
    \centering%
    \fboxsep=2.5mm
    \fboxrule=2pt
    \fcolorbox{lightgray}{white}{\parbox{#2}{%
        #3%
    }}%
    \IfBooleanTF{#1}{\end{figure*}}{\end{figure}}
}
\newtheorem{example}{Example}
\newtheorem{experiment}{Experiment}
\crefname{example}{Example}{Example}
\Crefname{example}{Example}{Examples}
\crefname{experiment}{Experiment}{Experiments}
\Crefname{experiment}{Experiment}{Experiments}
\crefname{figure}{Fig.}{Fig.}
\Crefname{figure}{Fig.}{Fig.}
\crefname{theorem}{theorem}{theorem}
\Crefname{theorem}{Theorem}{Theorems}
\crefname{lemma}{lemma}{lemma}
\Crefname{lemma}{Lemma}{Lemmas}
\crefname{definition}{definition}{definitions}
\Crefname{definition}{Definition}{Definitions}
\crefname{corollary}{corollary}{corollaries}
\Crefname{corollary}{Corollary}{Corollaries}
\crefname{proposition}{proposition}{propositions}
\Crefname{proposition}{Proposition}{Propositions}
\crefname{claim}{claim}{claims}
\Crefname{claim}{Claims}{Claims}
\crefname{problem}{problem}{problems}
\Crefname{problem}{Problem}{Problems}
\crefname{solution}{solution}{solutions}
\Crefname{solution}{Solution}{Solutions}
\crefname{proof}{proof}{proofs}
\Crefname{proof}{Proof}{Proofs}
\crefname{proofof}{proof}{proofs}
\Crefname{proofof}{Proof}{Proofs}

\newcommand{\warning}[1]{{\it\color{red}#1}}
\newcommand{\note}[1]{{\color{red}[Note: {\it #1}]}}
\newcommand{\nothing}[1]{}

\ifthenelse{\equal{\final}{1}}
{

\renewcommand{\warning}[1]{}
\renewcommand{\note}[1]{}
}
\newcommand{\eps}{\epsilon}

\newcommand{\sR}{\mathbb R}

\newcommand{\sC}{\mathbb C}

\newcommand{\e}{{\boldsymbol e}}

\newcommand{\s}{{\boldsymbol s}}

\newcommand{\x}{{\boldsymbol x}}
\newcommand{\y}{{\boldsymbol y}}

\newcommand{\A}{{\boldsymbol A}}
\newcommand{\B}{{\boldsymbol B}}

\newcommand{\D}{{\boldsymbol D}}

\newcommand{\F}{{\boldsymbol F}}

\newcommand{\I}{{\boldsymbol I}}

\newcommand{\K}{{\boldsymbol K}}
\renewcommand{\L}{{\boldsymbol L}}

\newcommand{\Q}{{\boldsymbol Q}}
\newcommand{\R}{{\boldsymbol R}}
\renewcommand{\S}{{\boldsymbol S}}

\newcommand{\U}{{\boldsymbol U}}
\newcommand{\V}{{\boldsymbol V}}
\newcommand{\W}{{\boldsymbol W}}
\newcommand{\X}{{\boldsymbol X}}
\newcommand{\Y}{{\boldsymbol Y}}
\newcommand{\Z}{{\boldsymbol Z}}
\newcommand\bmu{\boldsymbol{\mu}}
\newcommand\balpha{\boldsymbol{\alpha}}
\newcommand\bSigma{\boldsymbol{\Sigma}}

\newcommand\bGamma{\boldsymbol{\Gamma}}
\newcommand\bDelta{\boldsymbol{\Delta}}
\newcommand\bzero{\boldsymbol{0}}
\newcommand\bPhi{\boldsymbol{\Phi}}

\newcommand\dd{\mathop{}\!\operatorname{d}}

\DeclarePairedDelimiter\abs{\lvert}{\rvert} % requires mathtools package
\DeclarePairedDelimiter\norm{\lVert}{\rVert}
\newcommand{\grass}[2]{\operatorname{\mathcal{G}}\left( #1, #2 \right)}
\newcommand{\tgt}[1]{\operatorname{\mathcal{T_{#1}}}}

\newcommand{\spn}[1]{\operatorname{span}\left(#1\right)}
\newcommand{\rank}[1]{\operatorname{rank}\left(#1\right)}

\newcommand{\orth}[1]{\mathbf{O}\left(#1\right)}

\newcommand{\diag}[1]{ \operatorname{diag}\left(#1\right) }
\newcommand{\iprod}[2]{\left< #1, #2 \right>}

\newcommand{\bcd}[2]{\mathop{d_{\mathrm{BC}}}\left(#1, #2\right)} % Binet-Cauchy
\newcommand{\projd}[2]{\mathop{d_{\mathrm{P}}}\left(#1, #2\right)} % Projection
\newcommand{\fsd}[2]{\mathop{d_{\mathrm{FS}}}\left(#1, #2\right)} % Procrustes
\newcommand{\chd}[2]{\mathop{d_{\mathrm{C}}}\left(#1, #2\right)} % Similarity
\newcommand{\arcd}[2]{\mathop{d}\left(#1, #2\right)} % Geodesic metric

\newcommand{\mat}[1]{\boldsymbol{#1}}
\newcommand{\tp}[1]{#1 ^ {\top}}
\newcommand{\tr}[1]{\operatorname{tr}\left(#1\right)}

\ifthenelse{\equal{\final}{1}}
{
}

\begin{document}
\graphicspath{{sub/}{raster/}}
\title{Grassmannian Learning: Embedding Geometry Awareness in Shallow and Deep Learning}
%A Dichotomy of Leveraing Grassmann Manifolds in Shallow and Deep Models}
% Grassmaniann Learning: Towards Geometric Approach for Ad-Hoc Problems
% .. : Develoiping a geometric intruition towards better algorithm design
% .. : A geometric intruition towards
\date{}
\author{Jiayao~Zhang, Guangxu~Zhu, Robert~W.~Heath Jr.,
    and Kaibin~Huang
\thanks{{\small J. Zhang is with the Dept. of Computer Science, and
G. Zhu and K. Huang with the Dept. of Electrical \& Electronic Engr., all at The University of Hong Kong,
Hong Kong. R. W. Heath Jr. is with the Dept. of Electrical \& Computer Engr. at The University of Texas at Austin, TX, USA. Corresponding author: J. Zhang \texttt{<jiayaozhang@acm.org>}.}
}}
%\thanks{Manuscript received: }}

%\markboth{Preprint (Work in progress)}%
%{Zhang \MakeLowercase{\textit{et al.}}: Grassmannian Learning}

\maketitle

Modern machine learning algorithms have been adopted in a range of
signal-processing applications spanning computer vision, natural language
processing, and artificial intelligence. Many relevant problems involve
subspace-structured features, orthogonality constrained or  low-rank constrained objective
functions, or subspace distances. These mathematical characteristics are expressed
naturally using the Grassmann manifold. Unfortunately, this fact is not yet explored in
many traditional learning algorithms. In the last few years, there have been
growing interests in studying Grassmann manifold to tackle new learning problems.
Such attempts have been reassured by substantial
performance improvements in both classic learning and learning using deep neural
networks. We term the former as \emph{shallow} and the latter \emph{deep} Grassmannian learning.
The aim of this paper is to introduce the emerging area of Grassmannian learning by
surveying common mathematical problems and primary solution approaches, and
overviewing various applications. We hope to inspire practitioners in
different fields to adopt the powerful tool of Grassmannian learning in their
research.

% Modern machine learning
% algorithms have been widely adopted in a variety of applications in signal processing
% pertaining the core of computer vision, speech recognition, and artificial intelligence.
% Many of these problems involve subspace-structured features, orthogonality- or low-rank constrained
% objective functions, or subspace distances. These can be modelled naturally
% by the Grassmann manifold, yet are failed to be captured by many traditional learning algorithms.
% To that end, the utilization of the Grassmann manifold in classical learning problems has been placed under
% constant curiosity and has achieved commendable performance;
% novel methods incorporating state-of-the-art learning tools such as
% deep neural networks have been emerging and proven promising.
% We term the former as \emph{shallow} Grassmannian learning whilst the latter \emph{deep}
% Grassmannian learning.
% The aim of this paper is to review the notion and paradigms of Grassmannian learning,
% survey common problems and classical applications, overview various examples,
% and provide inspirations for practitioners in different fields
% with the Grassmannian learning toolbox.
%

%\begin{IEEEkeywords}
%Grassmann manifolds, machine learning, deep neural networks, computer vision, wireless communications
%\end{IEEEkeywords}

% INTRODUCTION
\section{Overview of Grassmann Learning}
\label{sec:introduction}
A Grassmann manifold refers to a space of subspaces embedded in a higher-dimensional vector space (e.g., the surface of a sphere in a 3D space).  The mathematical  notation  arises in a variety of science and engineering
    applications in the areas of  computer vision, statistical learning, wireless
    communications, and natural language processing. In visual recognition and classification tasks,
    the Grassmann manifold is used to model the invariant illumination or pose space
    \cite{Turaga:2008:SGCV,Turaga:2011:GSCV}. In statistical learning, novel methods
such as Grassmannian discriminant
    analysis \cite{Hamm:2008:GDA,Harandi:2011:GGDA,Souza:2016:EGDA} and clustering
\cite{Cetingul:2009:INMS,Wang:2017:GRSSC} are developed for processing  data on the Grassmann manifold or exploiting tools from Grassmannian optimization to enhance learning performannce.
    In recommender systems, under low-rank or sparsity constraints, the problem of matrix
    completion  can be solved using Grassmannian learning methods \cite{Dai:2012:GLRC,Absil:2015:LOWRANK}. In wireless communications, Grassmannian packing can be applied to  the design of
    space-time constellations
    \cite{Hochwald:2000:Unitary,Zheng:2002:GRASS,GrassMIMO} and that of limited feedback
    beamforming codebook  \cite{Love:2003:GB,Love:2005:LF}. In natural
    language processing, the Grassmann manifold can be used to model affine
    subspaces in document-specific language models \cite{Hall:2000:LCM}. These
    problems generally utilize the Grassmann manifold as a tool for nonlinear
    dimensionality reduction or to tackle optimization objectives that are
    \emph{invariant to the basis of domain}. This approach is not yet explored in other learning algorithms.
    The set of mentioned  problems belong to \emph{shallow learning} and will be surveyed in the first half of this paper.  In the second  half of the paper, we will discuss the latest trends in utilizing  Grassmann manifolds to exploit the geometry of problems for deep learning
    \cite{Lecun:2015:DL,Bronstein:2017:GEODL}. Relevant applications include shape
    alignment and retrieval \cite{Monti:2016:GEOCNN},  geodesic convolutional
    neural nets \cite{Masci:2015:GC}, and Riemannian
    curvature in neural networks \cite{Poole:2016:EXP}. Researchers have also proposed new
    deep neural network architectures for coping with data on the Grassmann manifold
    \cite{Huang:2016:GRNET,Herath:2017:ILS}.

    \begin{table*}[pbt]
    \centering
    \begin{tabularx}{\textwidth}{cXc}
    \toprule[1.5pt]
    Model & Learning Methods & Paradigm \\
    \midrule[1.5pt]
    \parbox[t]{1mm}{\multirow{3}{*}{\rotatebox[origin=c]{90}{Shallow}}} & Grassmannian Discriminant Analysis (GDA) \cite{Hamm:2008:GDA} (\Cref{sec:trad:disc}) & Kernel Method \\
    & Sparse Spectral Clustering (SSC) \cite{Wang:2017:GRSSC} (\Cref{sec:trad:ssc}) & Grassmannian Optimization \\
    & Low-rank Matrix Completion \cite{Dai:2012:GLRC} (\Cref{sec:trad:lrr}) & Grassmannian Optimization \\
    \midrule[1.5pt]
    \parbox[t]{2mm}{\multirow{3}{*}{\rotatebox[origin=c]{90}{Deep}}} & Sample Geodesic Flow (SGF) \cite{Gopalan:2011:GGD} with Deep Features (\Cref{sec:deep:tl}) & Grassmannian Optimization \\
 & Geodesic Flow Kernel (GFK) \cite{Gong:2012:GFK} with Deep Features (\Cref{sec:deep:tl}) & Kernel Method \\
 & Building deep neural nets on the Grassmann manifold \cite{Huang:2016:GRNET,Lohit:2017:LIR} (\Cref{sec:deep:net}) & Grassmannian Optimization \\ \midrule
    \bottomrule[1.5pt]
    \end{tabularx}
    \caption{Summary of representative Grassmannian learning methods.}
    \label{table:methods}
    \end{table*}

        \begin{table*}[pbt]
    \centering
    \begin{tabular}{cc}
    \toprule[1.5pt]
        Notation & Remark \\
    \midrule[1.5pt]
        $\sR^n, \sC^n$ & $n$-dimensional real and complex space \\
        $\mathcal{M}, \mathcal{H}$ & Arbitrary manifolds \\
        $\grass{n}{k}$ & $(n,k)$-Grassmann manifold \\
        $\orth{k}$ & Collection of $k\times k$ orthonormal (or unitary in the complex case) matrix \\
        $x, \x, \X$ & Scalar, vector, matrix or matrix representation of points on the Grassmann manifold \\
        $\tgt{\X}, \bDelta$ & Tangent space and tangent vector of $\mathcal{M}$ at $\X$ \\
        $\bPhi(\cdot)$ & Geodesic on the manifold \\
        $\exp$, $\log$ & Exponential and logarithm maps \\
        $\theta_i$ & Principal angle \\
        $F_{\X}$ & Matrix derivative of some function $F$ with respect to $\X$ \\
        $\nabla_{\X} F$ & Gradient of $F$ at point $\X$ \\
        $d(\cdot, \cdot)$ & A distance measure \\
        $k(\cdot, \cdot)$ & A kernel function \\
        CS Decomposition & Cosine-Sine Decomposition (\Cref{sec:pre:geo}) \\
        SPD & Symmetric Positive Definite (\Cref{sec:pre:dist}) \\
        RKHS & Reproducing Kernel Hilbert Space (\Cref{sec:pre:kernel}) \\ \midrule
            \bottomrule[1.5pt]
    \end{tabular}
    \caption{List of notations used in this paper.}
    \label{table:notations}
    \end{table*}

    The strength of Grassmannian learning lies in its capability to harness
    the structural information embedded in the problem, leading to lower complexity and improved  performance.
For example, the \emph{Grassmannian discriminant analysis} (GDA) applied to  image-set classification can better capture the subspace invariance of facial expressions than traditional methods do. As another example, in visual domain adaptation, \emph{Grassmannian geodesic flow kernel} (GFK) can exploit the domain-invariant features hidden in the geodesic (defined in Section \ref{sec:deep:tl}) connecting the source   and target domains, both being Grassmann manifolds,  to enable  effective knowledge transfer between them. More evident examples are provided in the sequel. As we observe,  the relevant techniques are scattered in diverse fields, and there lacks a systematic and accessible introduction to the Grassmann learning. The  existing introductory
work is either mathematically involved
    \cite{Edelman:1998:ALGO,Absil:2004:GEOM,Absil:2009:APPLE} or is documentation for software packages \cite{MANOPT,PYMANOPT}. For these reasons,  we  aim to provide an introduction to the Grassmannian manifold and its applications to both shallow and deep learning. To this end, two common paradigms in Grassmannian learning, namely, the \emph{Grassmannian kernel methods} and the \emph{Grassmannian optimization}, are introduced.  Representative applications in  these two paradigms are surveyed and summarized in \Cref{table:methods}.  We hope the discussion will facilitate readers in the signal processing community to tackle problems similar in nature.

\section{A Crash Course on Grassmann Manifolds}
\label{sec:preliminaries}

This section introduces notions from differential geometry, and then correlates them to the main theory.
A more comprehensive and rigorous treatment can be found in \cite{Boothby:1986:MANI,Wong:1967:DG}.
The notation used in this paper is summarized in \Cref{table:notations} for
ease of exposition.

\subsection{Definition of Grassmann Manifold} \label{sec:pre:def}

\begin{figure}[bt]
           \centering
           \def\svgwidth{0.7\linewidth}
           \input{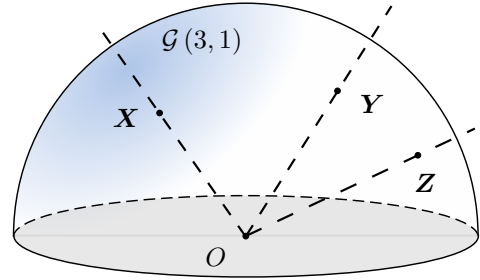}
           \caption{Example of a Grassmann manifold.
           The Grassmann manifold $\grass{3}{1}$ is the collection of lines in the Euclidean space $\sR^3$.
           The elements on $\grass{3}{1}$ are represented by points such as $\X$, $\Y$ and $\Z$,
       at the interceptions of the corresponding lines and the surface of the unit sphere in $\sR^3$.}
        \label{fig:ex_gr_31}
       \end{figure}

The Grassmann manifold $\grass{n}{k}$ with the integers $n \geq k > 0$ is the space
    formed by all $k$-dimensional linear subspaces embedded in an $n$-dimensional
    real or complex Euclidean space. As illustrated in \Cref{fig:ex_gr_31},
    the space of all lines passing through the origin in $\sR^2$ plane forms the
    real $\grass{3}{1}$.
Representation of elements on the Grassmann manifold is important for developing learning algorithms.
    An element on a Grassmann manifold is
    typically represented an arbitrarily chosen $n\times k$ orthonormal matrix $\X$
    whose column spans the corresponding subspace, called a \emph{generator} of the
    element. The Grassmann manifold can be represented by a
    collection of such generator matrices. Mathematically, this may be written as
    \begin{align}
        \grass{n}{k} = \left\{ \spn{\X} : \X \in \sR^{n \times k}, \tp{\X}\X = \I_k \right\}.
    \end{align}
    In our discussion, we will use $\X$ to represent a point on the Grassmann manifold [$\X \in \grass{n}{k}$],
    a subspace, or a orthonormal matrix ($\tp{\X}\X=\I$). The specific meaning will be clear in the context.
    Let $\orth{k}$ denote the collections of all $k\times k$
    orthonormal matrices. Since a subspace is represented by the span of the columns of $\X$,
    an element on the
    Grassmann manifold is invariant to rotations. Specifically, $\X$ and
    $\X\R$ correspond to the same point on $\grass{n}{k}$ for any $\R \in \orth{k}$.

       \subsection{Principal Angle} \label{sec:pre:pa}

        The distance between two elements $\X, \Y \in \grass{n}{k}$ on the Grassmann manifold is
        a function of the \emph{principal angels} $\{\theta_i\}_{i=1}^k$.
        The principal angles can be defined recursively by
        \begin{align} \label{eq:p_ang}
        \begin{cases}
        \cos \theta_{i} = \underset{\substack{\x_i\in\X
            \\\y_i\in\Y}} \max\,
            \tp{\x_i}\y_i\\
        \tp{\x_i}\x_i = 1, \quad \tp{\y_i}\y_i = 1, \\
        \tp{\x_i}\x_j = 0, \quad \tp{\y_i}\y_j = 0, \quad\quad \forall j < i,
        \end{cases}
        \end{align}
        for all $i=1,2,\ldots, k$.
        Intuitively, principal angles are the ``minimal'' angles between all possible bases
        of two subspaces.
        In practice, the principal angles between $\X, \Y \in \grass{n}{k}$
        can be computed from singular value decomposition (SVD),
        where the singular values of $\tp{\X}\Y$ are the cosines of the principal angles.
We will show  how different distance measures can be defined using the principal angles
	shortly in \Cref{sec:pre:dist}.

% \begin{figure*}
% \centering
% %\def\svgwidth{0.5\textwidth}
% \begin{subfigure}[t]{0.5\textwidth}
% \centering
% \def\svgwidth{0.9\textwidth}
% \input{01_R3_gr.tex}
% \caption{}\label{fig:prangle}
% \end{subfigure}%
% \begin{subfigure}[t]{0.5\textwidth}
% \def\svgwidth{0.9\textwidth}
% \centering
% \input{01_manifold.tex}
% \caption{}
% \label{fig:tangent}
% \end{subfigure}
% \caption{\emph{Left: Illustration of Principal Angles. }
% Bases of subspaces $\mathcal{U}$ and $\mathcal{V}$ are given.
% The $i$-th principal angle between them is thus the angel spanned by
% basis vectors $\u_i$ and $\v_i$, where they are the maximizer in
% \eqref{eq:p_ang}. They are also orthogonal to all previously defined
% basis vector in their span respectively.
% \emph{Right: Illustration of Tangent Space, Geodesic, Logarithm and Exponential Maps. }
% The two subspaces portrayed
% in the left can be viewed as two distinct points on some Grassmann manifold $\mathcal{M}$.
% The tangent space at $\X$ is a vector space $\mathcal{T}_{\X}\mathcal{M}$. The logarithm map
% maps the tangent vector to the point on the manifold whereas the exponential map does the
% reverse.}
% \end{figure*}

    \subsection{Tangent Space, Gradient and Retraction} \label{sec:pre:tgt}

        Gradient-based learning algorithms %such as steepest or conjugate gradient descent
        on the Grassmann manifold require the notion of tangency.
        For a point $\X \in \grass{n}{k}$, the
        space of tangent vectors $\tgt{\X}$ at $\X$ is defined as the set of all
        ``vectors" (matrices with a sense of direction to be precise) $\{\bDelta\}$ such that $\tp{\X} \bDelta = \bzero$.
        The gradient of some function $F: \grass{n}{k} \to \sR$ defined on a Grassmann
        manifold can be computed by projecting the ``Euclidean gradient''
        $\F_{\X} = \begin{bmatrix} \frac{\partial F}{\partial X_{ij}} \end{bmatrix}$
        onto the tangent space of the Grassmann manifold via the orthogonal projection
        $\F_{\X}\rightarrow \nabla_{\X} F$:
        \begin{equation} \label{eq:grad}
            \begin{aligned} \nabla_{\X} F = \left(\I_k
            - \X \tp{\X} \right) \F_{\X}. \end{aligned}
        \end{equation}
Since $(\I - \X\tp{\X})$ is an orthogonal projection  onto the orthogonal complement of $\X$,
        $\tp{\nabla F_{\X}}\X=\bzero$ and hence $\nabla F_{\X}$ is a tangent vector. The gradient computation in \eqref{eq:grad} plays an important role in Grassmannian
        optimization algorithms such as conjugate gradient descent. Such algorithms aim at finding the tangent matrix corresponding to the
        descent direction and computing a step forward on the manifold
        aligned in this direction. This requires an ``retraction" operation mapping a tangent matrix back onto the
        manifold through the exponential map, which will be defined explicitly after introducing the
        concept of geodesic.

        \subsection{Grassmann Geodesic} \label{sec:pre:geo}
	    Given two points on a manifold, a geodesic refers to the shortest curve on the
	    manifold connecting the points. Consider the earth as an example.
	    Mathematically, the earth surface is a manifold, namely a two dimensional
	    sphere embedded in the three dimensional Euclidean space, and geodesics are arcs
	    on the great circles of the earth. The trajectories of airliners or sea carriers
	    are conveniently represented as ``straight lines'' on a global map, but they
	    in fact travel on on great circles of the earth when viewed form the outer
	    space. Then the geodesic between a origin and a destination is the connecting
	    arc on a great circle passing the two points. Solving for geodesic is a
	    classical problem in the calculus of variation. In the case of
	    the Grassmann manifold, there exists relatively simple method of computing geodesics
	    using the relatively simple method based on the SVD \cite{Edelman:1998:ALGO,
	    Absil:2004:GEOM, Absil:2009:APPLE}.

        \textbf{Computing Grassmannian Geodesic: }
	    The geodesic between to points $\X, \Y \in \grass{n}{k}$ may be
	    parametrized by a  function $\bPhi(t):[0,1] \to \grass{n}{k}$, where
	    $\bPhi(0) = \X$ and $\bPhi(1) = \Y$. The parameter $t \in [0,1]$
	    controls the location on the geodesic and $t = \{0, 1\}$
	    corresponds to the two end points. To compute the geodesic on the
	    Grassmannian, consider the following operations. First, transport the point
	    $\X$ (a subspace) in the Euclidean space with the direction and
	    distance as specified by the tangent vector $\bDelta \in \tgt{\X}$.
Second, project the displaced point onto the manifold
	    $\grass{n}{k}$, giving the destination $\Y$. This particular
	    ``projection" operation is the exponential mapping mentioned in
	    Section \ref{sec:pre:tgt} and to be defined in the sequel. Note that
	    $\Y$ thus obtained is a subspace resulting from rotating $\X$ in the
	    direction $\bDelta$. Given the above operation and $\U \bSigma
	    \tp{\V}$ being the compact SVD of $\bDelta$, the Grassmann geodesic
	    between $\X$ and $\Y$ can be written as
	    \begin{align} \label{eq:grgeo}
            \bPhi(t) = \begin{bmatrix}
                    \X\V & \U
                        \end{bmatrix}
                        \begin{bmatrix}
                            \diag{\cos \bSigma t} \\
                            \diag{\sin \bSigma t}
                        \end{bmatrix}
                        \tp{\V}.
            \end{align}
	    where the sine and cosine act elementwisely on the diagonal of $\bSigma$ (i.e.,
	    the singular values of $\bDelta$). One can verify that $\X$ and $\Y$ are
	    two end points of the geodesic: $\X = \bPhi(0)$ and $\Y = \bPhi(1)$. Then the
	    exponential map, denoted as $\exp: \tgt{\X} \to \grass{n}{k}$, can be
	    defined as the computation of $\Y = \bPhi(1)$ using the origin $\X$ and
	    the tangent $\bDelta$.

	            \begin{figure}
        \centering
    % \begin{subfigure}[t]{0.5\textwidth}
    % \centering
    % \def\svgwidth{0.9\textwidth}
    % \input{01_R3_gr.tex}
    % \caption{}\label{fig:prangle}
    % \end{subfigure}%
    % \begin{subfigure}[t]{0.5\textwidth}
    % \def\svgwidth{0.9\textwidth}
    % \centering
    % \input{01_manifold.tex}
    % \caption{}
    % \label{fig:tangent}
    % \end{subfigure}
           \centering
           \def\svgwidth{0.9\linewidth}
           \input{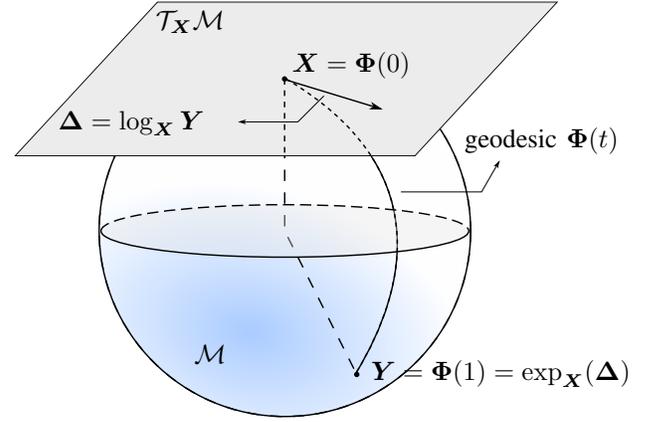}
        \caption{Consider two points $\X$ and $\Y$ on the Grassmann manifold represented by the sphere. The figure illustrates the tangent space at $\X$ denoted as $\mathcal{T}_{\X}\mathcal{M}$, the geodesic $\bPhi(t)$ connecting $\X$ and $\Y$, the logarithm map maps the tangent vector $\bDelta$ to the point on the manifold whereas the exponential map does the
	reverse.} \label{fig:grass}
        \end{figure}

	    \emph{How can we compute the Grassmann geodesic without the knowledge of the
	    tangent vector?}
	    In this case, the \emph{Cosine-Sine} (CS) decomposition is a tool that can
	    compute the vector $\bDelta$ pointing in the direction from $\X$, a point on the
	    Grassmannian $\grass{n}{k}$, to another $\Y$. For the two points $\X$ and
	    $\Y$, the CS decomposition can be viewed as the inverse
	    mapping of the exponential map from $\bDelta$ to $\Y$, which is thus
	    referred in the literature as
	    the \emph{logarithm map} $\log_{\X} \Y: \grass{n}{k} \to \tgt{\X}$.
	    Mathematically, the decomposition can be defined using the following equation:
	    \begin{equation} \label{eq:csd}
            \begin{aligned}
            \begin{bmatrix}
                \tp{\X} \Y \\
                (\I_n - \X \tp{\X})\Y \\
            \end{bmatrix}
            =
            \begin{bmatrix}
		\V \cos\left(\bSigma\right) \tp{\V} \\
		\U \sin\left(\bSigma\right) \tp{\V} \\
            \end{bmatrix},
            \end{aligned}
        \end{equation}
	    for some orthonormal matrices $\U \in \orth{n}$ and $\V \in
	    \orth{k}$. Then the logarithm map can be defined as $\log_{\X}: \Y \mapsto
	    \bDelta$ with $\bDelta = \U \bSigma \tp{\V}$. In practice, the CS decomposition
	    can be implemented based on the generalized SVD \cite{Edelman:1998:ALGO} which
	    computes the pair of SVD in \eqref{eq:csd}.
	    To summarize, we illustrate the quantities discussed above in
	    \Cref{fig:grass}. We will consider an example later
	    in \Cref{sec:deep} in the context of transfer learning.

        \subsection{Subspace Distance Measures} \label{sec:pre:dist}

	    Many machine learning algorithms require measuring the similarity
	    between data samples.
	    For example, in computer vision, the similarity between two images may
	    be measured by the sum of squared differences of each pixel, the
	    variation in the histogram of feature descriptors, the difference in
	    the latent representation, and many more. Similarly, in the applications
	    involving Grassmannian data, characterization of the discrepancy between
	    subspaces are usually needed. In the literature, many subspace-distance
	    measures have been defined and used, including the arc length
	    (corresponding to the geodesic distance) $d$, Fubini-Study distance
	    $d_{\mathrm{FS}}$, chordal distance $d_{\mathrm{C}}$, projection
	    distance $d_{\mathrm{P}}$, and Binet-Cauchy distance $d_{\mathrm{BD}}$.
	    In algorithmic design and analysis, a specific subspace distance measure
	    is chosen either for tractability or performance optimization. The
	    mathematical definitions of some commonly used measures are summarized
	    in \Cref{table:kernel}. The definitions reveal two ways of computing
	    the subspace distances between two points $\X$ and $\Y$ on the
	    Grassmannian: one is in terms of their principal angles
	    $\{\theta_i\}_{i=1}^k$ and the other based on the orthonormal matrices
	    $\X$ and $\Y$.
	    Furthermore, as shown in the table, the projection and Binet-Cauchy distances
	    also have their kernel-based definitions. More relevant details are given in the
	    subsequent discussion on kernel methods.

        \begin{table*}[pbt]
            \centering
            \begin{tabularx}{\textwidth}{ >{\raggedright} cXXc}
            \toprule[1.5pt]
            Metric & Principal Angle Formulation & Matrix Formulation & Kernel \\
            \midrule[1.5pt]
            Arc Length $d$ & $\left(\sum_{i=1}^k\theta_i^2\right)^{1/2}$ & -- & -- \\
                \midrule
            Fubini-Study $d_{\mathrm{FS}}$ & -- & $\arccos \abs{\det \tp{\X}\Y}$ & -- \\
                \midrule
            Chordal $d_{\mathrm{C}}$ & $2\left(\sum_{i=1}^k \sin^2 \frac{\theta_i}{2}\right)^{1/2}$ & $\norm{\X\U - \Y\V}_F$ & -- \\
            & $=\sqrt{2} \left(k - \sum_{i=1}^k \cos \theta_i\right)^{1/2}$ & $= \sqrt{2} \left(k - \tr{(\tp{\X}\Y\tp{\Y}\X)^{1/2}}\right)^{1/2}$ & \\
                \midrule
            Projection $d_{\mathrm{P}}$ & $\left(\sum_{i=1}^k \sin ^2 \theta_i \right)^{1/2}$ & $\norm{\X\tp{\X} - \Y\tp{\Y}}_F$ & $\norm{\tp{\X}\Y}^2_F$ \\
                \midrule
            Binet-Cauchy $d_{\mathrm{BC}}$ & $\left(1 - \prod_{i=1}^k \cos ^2 \theta_i \right)^{1/2}$ & -- & $\det (\tp{\X}\Y)^2$ \\
                \midrule
            \bottomrule[1.5pt]
            \end{tabularx}
	    \caption{Several common distance measures between two points $\X$ and $\Y$ on the Grassmann manifold where $\{\theta_i\}_{i=1}^k$ are principal angles between $\X$ and $\Y$. }
            \label{table:kernel}
        \end{table*}

	The subtle differences between various measures can be explained intuitively as
	follows. The arc length is the length of the Grassmann geodesic and the
	longest among all distances. The chordal and projection distances both involve
	embedding the Grassmann manifold in higher dimensional Euclidean spaces and
	consider the familiar
	F-norm therein (other norms such as $2$-norms may also be used, which
	leads to e.g., projection $2$-norm). For example, the chordal distance
	embeds the Grassmann manifold $\grass{n}{k}$ in the $(n \times
	k)$-dimensional Euclidean space while the projection distance embeds
	$\grass{n}{k}$ in the $n \times n$ Symmetric Positive-Definite (SPD)
	manifold, formed by real $n\times n$ SPD matrices. A distance defined in
	a higher dimensional ambient space tends to be \emph{shorter} since
	``cutting a shorter path'' is possible. For example, a chord is short
	than an arc between the same two points. Mathematically, we have the
	following inequalities among several distance measures
	\cite{Edelman:1998:ALGO}:
        for any $\X, \Y \in \grass{n}{k}$,
        \begin{equation} \label{eq:dists}
        \arcd{\X}{\Y} > \chd{\X}{\Y} > \projd{\X}{\Y}, \\
        \arcd{\X}{\Y} > \fsd{\X}{\Y}.
        \end{equation}
	Note the chordal distance can be rewritten as $d_{\mathrm{C}} =
	\sqrt{2}\left(\sum_{i=1}^k \sin^2 \frac{\theta_i}{2}\right)^{1/2}$. It
	is worth mentioning that removing $\sqrt{2}$ in the above expression
	gives another distance measure, the \emph{Procrustes distance},
	frequently used in shape analysis \cite{Chikuse:2012:STAT}. For
	illustration, we provide two examples in \Cref{ex:dist}.

    \FloatBox*[\linewidth]{
        \begin{example}[Subspace distances] \label{ex:dist} \normalfont
	Two concrete examples of computing distances between two points on a Grassmann
	manifold are given as follows.
        \\
	$\spadesuit$ \; { \bf Left:} As a simple example, consider two
	points $\X = \tp{[ 1 \quad 0 ]}$ and $\Y = \tp{ [ \frac{1}{2} \quad
	\frac{\sqrt{3}}{2} ]}$ on the Grassmannian $\grass{2}{1}$. They
	have only a single principal angle of $\frac{\pi}{3}$. Based on
	Table \ref{table:kernel} and illustrated in Fig. \ref{fig:sub_dis}, the
	arc length between $\X$ and $\Y$ is the length of the geodesic joining
	the points, namely $d = \frac{\pi}{3}$; the chordal distance is the
	chord joining them, computed as $d_{\mathrm{C}} = 1$; the Projection
	distance is the length of the projection from $\X$ to $\Y$ is $d_P = \frac{\sqrt{3}}{2}$.
        \\
	$\spadesuit$ \; {\bf Right:} As a more general example, consider
            \begin{align*}
                \X = \begin{bmatrix}
                        -\frac{\sqrt{2}}{2} & -\frac{\sqrt{2}}{4} \\
                        \frac{\sqrt{2}}{2} & \frac{\sqrt{2}}{4} \\
                        0 & \frac{\sqrt{3}}{2}
                    \end{bmatrix},\quad
                \Y = \begin{bmatrix}
                        0 & \frac{\sqrt{2}}{2} \\
                        1 & 0 \\
                        0 & \frac{\sqrt{2}}{2} \\
                    \end{bmatrix}
            \end{align*}
	as two points on the Grassmannian $\grass{3}{2}$. From the SVD of
	$\tp{\X}\Y$,
	the singular values are computed as $1.0$ and $0.079$. It follows from the
	results and Section \ref{sec:pre:pa} that the principal angles between $\X$ and
	$\Y$ are
        $\theta_1 = 0$, $\theta_2 = \arccos (0.07945931) \approx 85.44 \deg$.
	Using \Cref{table:kernel}, different subspace distances between the
	points are computed as $\arcd{\X}{\Y}\approx 1.491253$,
        $\fsd{\X}{\Y}\approx 1.491253$, $\chd{\X}{\Y}\approx 1.356864$,
        $\projd{\X}{\Y}\approx 0.996838$ and $\bcd{\X}{\Y}\approx 0.996838$,
        which confirm the relation in \eqref{eq:dists}. \\
        \begin{center}
            \begin{minipage}[c]{0.5\linewidth}
            \def\svgwidth{0.8\linewidth}
            \hspace*{2mm}
            %% Creator: Inkscape inkscape 0.92.2, www.inkscape.org
%% Accompanies image file '02_metric.pdf' (pdf, eps, ps)
\begingroup%
  \makeatletter%
  \providecommand\color[2][]{%
    \errmessage{(Inkscape) Color is used for the text in Inkscape, but the package 'color.sty' is not loaded}%
    \renewcommand\color[2][]{}%
  }%
  \providecommand\transparent[1]{%
    \errmessage{(Inkscape) Transparency is used (non-zero) for the text in Inkscape, but the package 'transparent.sty' is not loaded}%
    \renewcommand\transparent[1]{}%
  }%
  \providecommand\rotatebox[2]{#2}%
  \ifx\svgwidth\undefined%
    \setlength{\unitlength}{696.71445873bp}%
    \ifx\svgscale\undefined%
      \relax%
    \else%
      \setlength{\unitlength}{\unitlength * \real{\svgscale}}%
    \fi%
  \else%
    \setlength{\unitlength}{\svgwidth}%
  \fi%
  \global\let\svgwidth\undefined%
  \global\let\svgscale\undefined%
  \makeatother%
  \begin{picture}(1,0.75881169)%
    \put(0,0){\includegraphics[width=\unitlength,page=1]{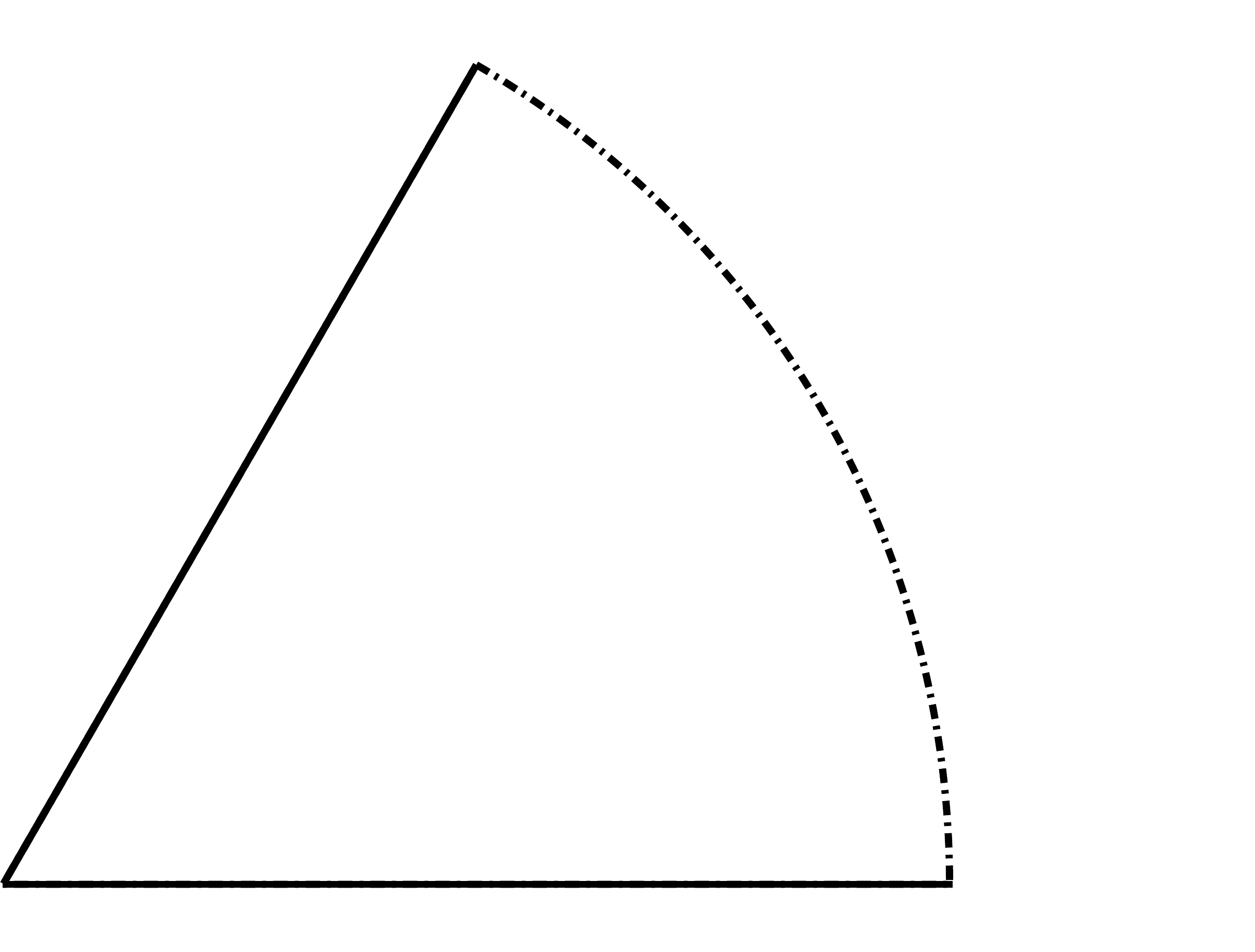}}%
    \put(0.38148859,0.74136372){\color[rgb]{0,0,0}\makebox(0,0)[lb]{\smash{$\boldsymbol{Y}$}}}%
    \put(0.7874286,0.07192981){\color[rgb]{0,0,0}\makebox(0,0)[lb]{\smash{$\boldsymbol{X}$}}}%
    \put(0,0){\includegraphics[width=\unitlength,page=2]{02_metric.pdf}}%
    \put(0.12766739,0.16454072){\color[rgb]{0,0,0}\makebox(0,0)[lt]{\begin{minipage}{0.29098633\unitlength}\raggedright {\small $\theta=\frac{\pi}{3}$}\end{minipage}}}%
    \put(0.00189141,0.04090622){\color[rgb]{0,0,0}\makebox(0,0)[lt]{\begin{minipage}{0.1967901\unitlength}\raggedright {\small $(0,0)$}\end{minipage}}}%
    \put(0.76047428,0.04090622){\color[rgb]{0,0,0}\makebox(0,0)[lt]{\begin{minipage}{0.1967901\unitlength}\raggedright {\small $(0,1)$}\end{minipage}}}%
    \put(0.66987753,0.54850813){\color[rgb]{0,0,0}\makebox(0,0)[lt]{\begin{minipage}{0.2640743\unitlength}\raggedright {\small $d=\frac{\pi}{3}$}\end{minipage}}}%
    \put(0.63347901,0.32900687){\color[rgb]{0,0,0}\makebox(0,0)[lt]{\begin{minipage}{0.2640743\unitlength}\raggedright {\small $d_{\mathrm{C}}=1$}\end{minipage}}}%
    \put(0.39715774,0.16414371){\color[rgb]{0,0,0}\makebox(0,0)[lt]{\begin{minipage}{0.2640743\unitlength}\raggedright {\small $d_{\mathrm{P}}=\frac{\sqrt{3}}{2}$}\end{minipage}}}%
  \end{picture}%
\endgroup%

        \end{minipage}%
        \begin{minipage}[c]{0.5\linewidth}
            \def\svgwidth{0.8\linewidth}
        %% Creator: Inkscape inkscape 0.92.2, www.inkscape.org
%% Accompanies image file '02_gr_31.pdf' (pdf, eps, ps)
\begingroup%
  \makeatletter%
  \providecommand\color[2][]{%
    \errmessage{(Inkscape) Color is used for the text in Inkscape, but the package 'color.sty' is not loaded}%
    \renewcommand\color[2][]{}%
  }%
  \providecommand\transparent[1]{%
    \errmessage{(Inkscape) Transparency is used (non-zero) for the text in Inkscape, but the package 'transparent.sty' is not loaded}%
    \renewcommand\transparent[1]{}%
  }%
  \providecommand\rotatebox[2]{#2}%
  \ifx\svgwidth\undefined%
    \setlength{\unitlength}{526.43250618bp}%
    \ifx\svgscale\undefined%
      \relax%
    \else%
      \setlength{\unitlength}{\unitlength * \real{\svgscale}}%
    \fi%
  \else%
    \setlength{\unitlength}{\svgwidth}%
  \fi%
  \global\let\svgwidth\undefined%
  \global\let\svgscale\undefined%
  \makeatother%
  \begin{picture}(1,0.85597318)%
    \put(0,0){\includegraphics[width=\unitlength,page=1]{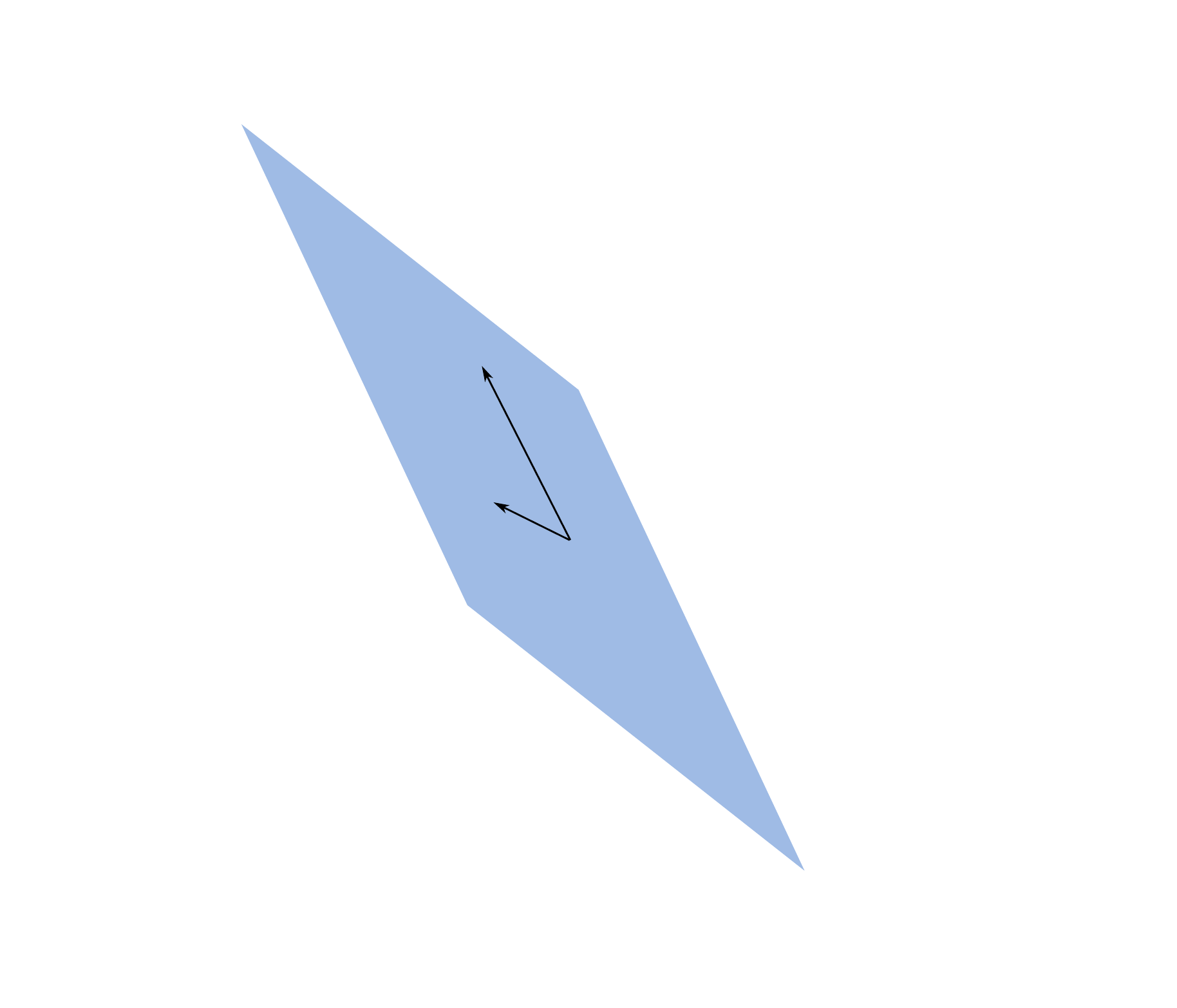}}%
    \put(0.43367324,0.82093489){\color[rgb]{0,0,0}\makebox(0,0)[lb]{\smash{$z$}}}%
    \put(0.9289884,0.35862489){\color[rgb]{0,0,0}\makebox(0,0)[lb]{\smash{$x$}}}%
    \put(0.94584716,0.48186006){\color[rgb]{0,0,0}\makebox(0,0)[lb]{\smash{$y$}}}%
    \put(0.27529586,0.72714319){\color[rgb]{0,0,0}\makebox(0,0)[lb]{\smash{$\X$}}}%
    \put(0,0){\includegraphics[width=\unitlength,page=2]{02_gr_31.pdf}}%
    \put(0.83809059,0.64783578){\color[rgb]{0,0,0}\makebox(0,0)[rb]{\smash{$\Y$}}}%
    \put(0.39883225,0.56086132){\color[rgb]{0,0,0}\makebox(0,0)[rb]{\smash{$\x_2$}}}%
    \put(0.35455857,0.42120724){\color[rgb]{0,0,0}\makebox(0,0)[lb]{\smash{$\x_1$}}}%
    \put(0.59927788,0.49860947){\color[rgb]{0,0,0}\makebox(0,0)[lb]{\smash{$\y_2$}}}%
    \put(0.61365833,0.40468967){\color[rgb]{0,0,0}\makebox(0,0)[lb]{\smash{$\y_1$}}}%
    \put(0,0){\includegraphics[width=\unitlength,page=3]{02_gr_31.pdf}}%
    \put(0.50063339,0.35102657){\color[rgb]{0,0,0}\makebox(0,0)[lb]{\smash{$O$}}}%
  \end{picture}%
\endgroup%
\label{fig:sub_dis}
        \end{minipage}
	\caption*{\textbf{Example: Geometric illustrations of subspace distance measures.}}
        \vspace{-6mm}
        \end{center}
        \end{example}
    }

        \subsection{Grassmann Kernel Methods} \label{sec:pre:kernel}

        \subsubsection{Background on Kernel Methods}
	    In many machine learning applications, to better explore the latent data
	    structure, a common practice is to project data into some high-dimensional
	    feature space through a specific mapping and train the model there. It is
	    expected that the low-dimensional data can be better disentangled in the higher
	    dimension where the data structure is more clear.
	    % XXX: Add 1 sentence on why (more differential) XXX
	    However, such training involves the computation of the coordinates of the projected data samples and their pairwise distances all in the high-dimensional feature space, resulting in high computation complexity.

	    The kernel method overcomes the difficulties by introducing a  kernel function
	    $k(\cdot,\cdot)$ associated with a corresponding mapping $\phi(\cdot)$. This pair
	    of functions induces a specific high-dimensional feature space.
A Kernel method allows us to efficiently compute the similarity  between two  data samples
	    in the feature space without the need to compute $\phi(\cdot)$ explicitly,
which is in general  difficult and in some cases intractable. The mathematical principle
	    of kernel methods are as follows. The kernel function $k(\cdot,\cdot)$ and the
	    mapping $\phi(\cdot)$ uniquely determine a \emph{reproducing kernel
	    Hilbert space} (RKHS), which is a vector space endowed with a proper inner product,
	    denoted as $\iprod{\cdot}{\cdot}_{\mathcal{H}}$, and satisfying a reproducing
	    property. Mathematically,
	    $k(\x,\y)=\iprod{\x}{\y}_{\mathcal{H}}=\tp{\phi(\x)}\phi(\y)$. Exploiting the
	    property, the dimensionality-sensitive operation of inner-product involved in
	    the distance evaluation in the high-dimension feature space, denoted as
	    $d_{\mathcal{H}}^2(\cdot,\cdot)$, can be replaced by the evaluation of the
	    computationally-friendly kernel function. This exploits the following
	    mathematical relation:
		\begin{align} \label{eq:kernel_dist}
		   d_{\mathcal{H}}^2(\x,\y) &= \tp{\phi(\x)}\phi(\x) + \tp{\phi(\y)}\phi(\y) - 2\tp{\phi(\x)}\phi(\y) \notag \\
				   &=k(\x,\x)+k(\y,\y)-2k(\x,\y).
		\end{align}
	     Note that for the inner-product to be properly defined, the
	     kernel function $k(\cdot, \cdot)$ should be symmetric
	     ($k(\x,\y)=k(\y,\x)$) and positive-definite ($k(\x,\y) > 0$ for all
	     $\x,\y \ne \bzero$).

	    Besides computational efficiency, kernel methods have gained their popularity in
	    learning also for other advantages including the existence of a wide range of
	    kernels and their capability of dealing with infinite-dimensional feature
	    spaces. Consider the Gaussian kernel as an example.
	    Given two data samples $x, y\in\sR$, the Gaussian kernel $k(x,y)
	    = \exp\left\{-\frac{1}{2\sigma^2} \norm{x-y}^2\right\}$ can be
	    evaluated. The kernel expression implicitly defines
        an \emph{infinite-dimensional} feature space inducted by the following mapping:
        \begin{align}
       \phi(x) = \exp\left\{-x^2/2\sigma^2\right\}
		\begin{bmatrix} 1, \sqrt{\frac{1}{1! \sigma^2}}x, \ldots, \sqrt{\frac{1}{n! \sigma^{2n}}} x^n, \ldots\end{bmatrix}^T.
        \end{align}

      \subsubsection{Grassmannian Kernel and Learning}
	Learning from data sets with elements being subspaces (e.g., image features or
	motions) has motivated the development of the Grassmann kernel methods. Simply
	by defining kernel functions on the Grassmann manifold $k(\cdot, \cdot):
	\grass{n}{k} \times \grass{n}{k}\to \sR^*$ and kernel replacement, classic
	kernelized learning algorithms in the Euclidean space can be readily migrated
	onto the Grassmann manifold. The key property distinguishing a Grassmann kernel
	from others is that the kernel function must be invariant to the choice of
	specific basis in subspace data samples. In other words,
	given two points $\X,\Y$ on some Grassmann manifold, the Grassmann
	kernel $k\left( \X, \Y \right) = k\left( \X\U, \Y\V \right)$ for any
	$\U, \V \in \orth{O}$. Two commonly used Grassmann kernels are the
	Binet-Cauchy kernel and the projection kernel given
	Table~\ref{table:kernel}.
	Similar to Euclidean kernels, the sum, product and composition of Grassmannian
	kernels also result in valid Grassmannian kernels. A detailed treatment of
	Grassmannian kernels can be found in e.g., \cite{Harandi:2014:EGK}.

\begin{figure}
        \centering
    \resizebox{\linewidth}{!}{%
        \input{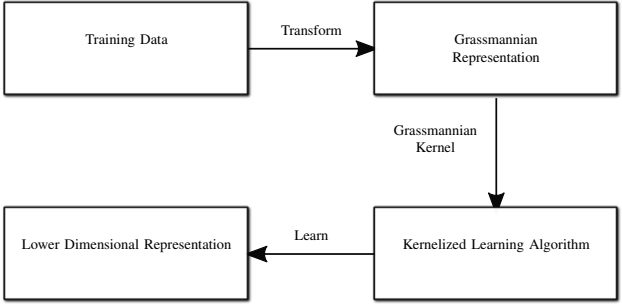}
    }
        %\resizebox{0.6\textwidth}{!}{%
        %\def\svgwidth{\linewidth}
        %\input{grass_op_paradigm.tex}}
        \caption{General framework for Grassmannian kernelized learning.}
        \label{fig:grass_data:paradigm}
    \end{figure}

	The general framework of applying a Grassmann kernel method in
	learning from Grassmannian data is shown in
	\cref{fig:grass_data:paradigm}. Raw data such as an image set is
	usually transformed to subspace features or Grassmannian representations
	using, for example, \emph{principle component analysis} (PCA). By
	choosing a  specific Grassmannian kernel function, the Grassmannian data
	can be fed into a kernelized learning algorithm such as classification
	based on support vector machine or linear discriminant analysis. We will
	revisit Grassmann kernelized learning in the discussion of linear
	discriminant analysis in \Cref{sec:trad:disc} and in deep transfer
	learning in \Cref{sec:deep:tl}.

% Besides the arc length, several other characterizations of the
% distance on the Grassmann manifold (also known as the discrepancy between subspaces)
% exist and are useful under different scenarios. Some of the distances may be induced
% by kernel functions on the Grassmann manifold. To be a valid \emph{Grassmann Kernel},
% besides symmetricity and positive-semidefiniteness, the invariance of representations
% among basis \cite{Hamm:2009:EGK} is required, namely,
% for any $\Y_i$ such that $\spn{\Y_1} = \spn{\Y_2}$ and $\spn{\Y_3} = \spn{\Y_4}$.
% Several other regularization conditions are also needed, one may refer
% to \cite{Hamm:2009:EGK} for further details.
%
% \begin{definition}[Grassmannian Kernel] \label{def:grass_kernel}
% Function $k: \grass{n}{k} \times \grass{n}{k} \mapsto \sR^*$
% is a Grassmannian kernel if and only if the following holds
% simultaneously:
% \end{definition}
%
% Indeed, performing proper arithmetic such as addition, exponentiation and composition of kernels
% also yield valid kernels \cite{Bishop:2006:PRML}. Interested readers are referred to
% \cite{Harandi:2014:KERNELS} for proper generalizations and comparisons of kernels induced from
% Cauchy-Binet and projection kernels.
%
% There exists a variety of different metrics defined on the Grassmann manifold,
% and there is indeed inconsistency in the naming convention of these metrics
% in the literature. Here we shall adopt the definition given by \citeauthor{Edelman:1998:ALGO} \cite{Edelman:1998:ALGO}
%

    \subsection{Optimization on Grassmann Manifolds} \label{sec:pre:op}

	Grassmannian kernel methods provide a tool for solving the class of problems
	involving Grassmannian data, or called Grassmann kernelized learning.
There also exists another class of problems that involve
	optimizing variables under specific structural constraints such as sparsity or
	low rank in the context of matrix completion. They can be often cast as
	optimization problems on the Grassmann manifold, which has the typical form in
	\eqref{eq:grassprob} or its equivalence in \eqref{eq:eqgrassprob}. Solving such
	a problem represents a search for a subspace or
	orthogonality-constrained optimization. One example is low-rank matrix
	completion in \Cref{sec:trad:lrr} where the goal is to find a subspace
	that is consistent with the observed entries. Another example is the method
	of sample geodesic flow method for
	transfer learning in \Cref{sec:deep:tl}, which yields a  subspace where data appears most discriminative.

	Grassmann optimization problem are usually solved using is gradient-based
	methods such as steepest or conjugate gradient descent on the Grassmann
	manifold. Compared with their Euclidean-space counterparts, the key feature of such
	methods is the computation of a gradient on the Grassmannian manifold using the
	formula in \eqref{eq:grad}. As discussed earlier, the gradient computation
	using \eqref{eq:grad} has low complexity which first evaluates the ``Euclidean
	gradient'' and then projects it onto the Grassmannian manifold to obtain the
	Grassmannian gradient. In practice, software packages such as
	\texttt{ManOpt} \cite{MANOPT} are available for Grassmann gradient computation.
	In addition, there exist problem-specific methods for Grassmannian optimization
	such as convex relaxation in the sparse spectral clustering. We will apply
	Grassmannian optimization methods to sparse and low-rank representation
	learning in \Cref{sec:trad:lrr} and deep learning in \Cref{sec:deep:net}.

    \begin{empheq}[box=\widefbox]{equation} \label{eq:grassprob}
        \begin{alignedat}{2}
        \min \quad & f \left( \X \right), \\
        \text{s.t.} \quad & \X \in \grass{n}{k}.
        \end{alignedat}
    \end{empheq}

    \begin{empheq}[box=\widefbox]{equation} \label{eq:eqgrassprob}
        \begin{alignedat}{2}
        \min \quad& f(\X), \\
        \text{s.t} \quad & f(\X) = f(\X \S),\\
        \text{where} \quad& \X \in \sR^{n \times k}, \S \in \orth{k}.
        \end{alignedat}
    \end{empheq}

    In summary, Grassmannian kernelized methods and Grassmannian optimization
    are two problem-solving paradigms targeting two different types of
    problems in Grassmannian learning. Their main differences are summarized
    in Table~ \ref{table:diff}.

    \begin{table}[htpb]
        \centering
	\begin{tabular*}{\linewidth}{ccc}
        \toprule[1.5pt]
        Paradigm & Input Data & Optimization Domain \\ \midrule[1.5pt]
        Grassmannian Kernel Methods & Grassmannian & RKHS \\ \midrule
        Grassmannian Optimization & General & Grassmannian \\ \midrule
        \bottomrule[1.5pt]
        \end{tabular*}
        \caption{Comparison between two learning paradigms: Grassmannian kernel methods and Grassmannian optimization.}
        \label{table:diff}
    \end{table}

\section{Shallow Grassmannian Learning}
\label{sec:traditional}

    This section is devoted to shallow Grassmannian learning methods, where
    Grassmann manifolds provide a tool for nonlinear dimensionality reduction.
% Reduced-dimension learning by exploiting sparsity is a recurring theme in many
% real-word applications pertinent to signal processing. For instance, sparsity
% may result from missing data in observations such as an unknown movie rating by
% a user or the inherent structure or underlying assumptions of a model. Many
% learning algorithms leveraging sparsity, such as
% Lasso, have been developed and their effectiveness are repeatedly proved in
% practice \cite{Hastie:2015:SPARSE}.
    In this section, we review several applications of shallow Grassmannian learning, including discriminant analysis in
    \Cref{sec:trad:disc}, high-dimensional data clustering in
    \Cref{sec:trad:ssc}, and low-rank matrix completion in
    \Cref{sec:trad:lrr}. The problems share a common theme of
    dimensionality reduction. However, their goals in representation differ.
    Discriminant analysis seeks a low-dimensional subspace where data are most
    discriminative; high-dimensional data clustering and low-rank matrix
    completion attempt to learn sparse and low-rank representations. Inline
    with traditional dimensionality-reduction techniques that directly operate
    on Euclidean data, with proper notion
    of Grassmann kernels and distance measures as introduced in
    \Cref{sec:preliminaries}, problems involving subspace data or operations on
    Grassmann manifolds may be made tractable or tackled more efficiently without
    sacrificing the geometric intuitions. For a more comprehensive and in-depth
    treatment of the topics, readers are referred to
    \cite{Hastie:2015:SPARSE,Cunningham:2015:LDR}.

\subsection{Grassmann Discriminant Analysis}
\label{sec:trad:disc}
       \begin{figure*}[bt]
           \centering
           \def\svgwidth{0.8\linewidth}
           \input{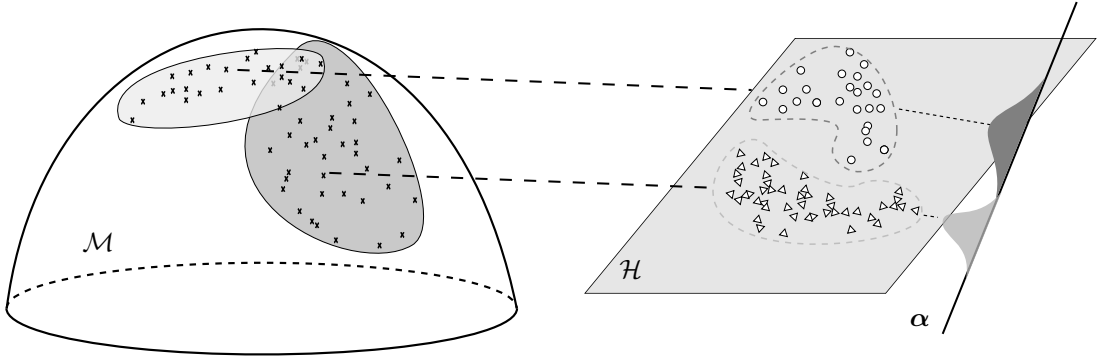}
       \caption{Illustration of kernelized GDA:
       %Data on the Grassmann manifold $\mathcal{M}$ is first projected into some RKHS $\mathcal{H}$.
Kernelized GDA firstly projects data on the Grassmann manifold $\mathcal{M}$
            to some RKHS $\mathcal{H}$
           constructed from a specific  Grassmannian kernel, and then performs
           discriminant analysis in  this space.}
            \label{fig:gda_proj}
       \end{figure*}
GDA builds on \emph{linear
    discriminant analysis} (LDA). The latter is a class of supervised learning
    algorithms for classification based on identifying a latent subspace in which
    the data are most discriminative, meaning that similar data samples are close
    but distant from dissimilar samples \cite{Bishop:2006:PRML}. The mathematical
    principle of LDA is described as follows. To this end, let $N$, $C$, $N_c$,
    $\bmu$, $\bmu_c$, $\x_i$ denote the dataset size, number of data classes, $c$-th
    class size, population mean, class mean and the $i$-th data sample.
    Considering class $c$, the \emph{intra-class covariance} with
    respect to the class mean $\bmu_c$ is
    \begin{equation} \label{eq:trad:disc:sw}
        \S_w = \frac{1}{N} \sum_{c=1}^C \sum_{i:y_i=c}
	\left( \x_i - \bmu_c \right) \tp{ \left( \x_i - \bmu_c \right) },
    \end{equation}
    while  the \emph{inter-class covariance} is defined in terms of population and class means as
    \begin{equation} \label{eq:trad:disc:sb}
        \S_b = \frac{1}{N} \sum_{c=1}^C N_c \left( \bmu_c - \bmu \right)
        \tp{ \left( \bmu_c - \bmu \right) }.
    \end{equation}
    Then finding the latent discriminant subspace can be translated into the concrete
    problem of finding a low-dimensional subspace, denoted as $\tp{\balpha}$, in which the projected
    inter-class covariance is maximized and the projected intra-class covariance
    minimized. The two objectives can be combined, leading to the optimization of
    the Rayleigh quotient
    \begin{equation} \label{eq:robj}
        f(\balpha) = \frac{\tp{\balpha} \S_b \balpha}{\tp{\balpha} \S_w \balpha}.
    \end{equation}

Based on the covariance definition in \eqref{eq:trad:disc:sw} and \eqref{eq:trad:disc:sb},
    LDA targets generic  data distributed in the linear
    Euclidean space.
    For applications such as image-set classification or action recognition, direct
    extension of LDA to handle Grassmannian data, meaning data on a nonlinear
    Grassmann manifold, is not trivial. First of all, how should mean and variance
    be defined on the manifold? As the manifold is nonlinear and data similarity is
    measured using subspace distances, the arithmetic in \eqref{eq:trad:disc:sb} and
    \eqref{eq:trad:disc:sw} are no longer meaningful and need be modified. There do exist a
    handful of notions of mean on the Grassmann manifold such as Procrustes mean and
    Karcher mean \cite{Chikuse:2012:STAT}. Their computation, however,
    typically involves solving a Grassmannian-optimization problem by a iterative
    Grassmannian-gradient method. As first proposed in \cite{Baudat:2000:KDA}, a
    more efficient and systematic approach for GDA is to develop kernelized
    discriminant analysis and leverage Grassmann kernels introduced in Section
    \ref{sec:pre:kernel}.

    The essence of the Grassmannian kernel approach is to define the Grassmann
    counterparts of the covariance matrices in \eqref{eq:trad:disc:sb} and
    \eqref{eq:trad:disc:sw} in terms of a kernel matrix. Let $\K$ denote such a
    matrix where the element $\K_{ij} = k(\X_i, \X_j)$ for some Grassmannian kernel
    function $k(\cdot, \cdot)$ defined in \Cref{sec:pre:kernel}
    and two Grassmann data samples $\X_i$ and $\X_j$.
Define $\begin{bmatrix}\frac{1}{N}\end{bmatrix}$ as an $N\times N$ matrix with each
    entry being $1/N$ and $\V$ as an $N$ by $N$ block diagonal matrix with the
    $i$-th block being $\begin{bmatrix}\frac{1}{N_C}\end{bmatrix}$,
    the intra-class and inter-class covariance matrices for GDA can be written as
    $\S_w=\K\left(\I_N - \V \right)\K$ and $\S_b =\K\left(\V-\begin{bmatrix}\frac{1}{N}\end{bmatrix}\right)\K$
    \cite{Hamm:2008:GDA}. The definitions allow the quotient minimization in
    \eqref{eq:robj} to be modified to a kernelized version for GDA as
     \cite{Hamm:2008:GDA}
        \begin{equation} \label{eq:gda_rq}
        \begin{split}
        f \left( \mat{\alpha} \right)
        &= \frac{ \tp{\mat{\alpha}} \mat{K} \left( \mat{V} - [\frac{1}{N}] \right) \mat{K}
            \mat{\alpha} }{ \tp{\mat{\alpha}} \left( \mat{K} \left(\mat{I}_N -  \mat{V}\right)
            \mat{K} + \eps^ 2 \mat{I}_N \right)\mat{\alpha} },
        \end{split}
        \end{equation}
	where $\eps^2 \I$ is optional and  added for numerical robustness in practice.

    The GDA is an exemplar application of the general kernel methods on the
    Grassmann manifold discussed in Section \ref{sec:pre:kernel}. The learning
    process is illustrated in \cref{fig:gda_proj} for a dataset consisting of two
    classes. The Grassmannian data (obtained by preprocessing the raw data, for
    example) can be viewed as being projected to an RKHS implicitly defined by the
    kernel function where data discriminative properties are retained. Thereby, LDA
    algorithms can be applied in this RKHS governed by a distance measure induced by
    the kernel. This yields a low-dimensional subspace where a classifier for
    Grassmannian data can be trained and subsequently applied to label future new
    data. By substituting proper Grassmannian kernels, learning on the Grassmann
    manifold can be built on top of the Euclidean kernelized learning algorithm with
    ease. We will revisit the GDA approach in the context of image classification in
    Section~\ref{sec:applications}.

    %Given the kernelized formulation of GDA in \eqref{eq:gda_rq}, a natural
    %question to ask is how to solve such a problem. A powerful approach is to apply
    %Grassmann kernel methods discussed in Section \ref{sec:pre:kernel} to tap into
    %the rich literature of LDA algorithms (see e.g., [XXX]). To be specific, the
    %Grassmannian data can be projected into a RKHS using a Grassmann kernel
    %function such that the data discriminative properties are retained. Thereby,
    %LDA algorithms can be applied in this RKHS governed by a distance measure
    %induced by the kernel. This yields a low-dimensional subspace where a
    %classifier for Grassmannian data can be trained and subsequently applied to
    %label future new data. The above learning process is illustrated in
    %\cref{fig:gda_proj}
    %for the special case of a dataset consisting of two classes. We will revisit the GDA approach in the context of image classification in Section~\ref{sec:applications}.

\subsection{High-Dimensional Data Clustering}
\label{sec:trad:ssc}
\begin{figure}[t!]
        \centering
    \resizebox{\linewidth}{!}{%
        \input{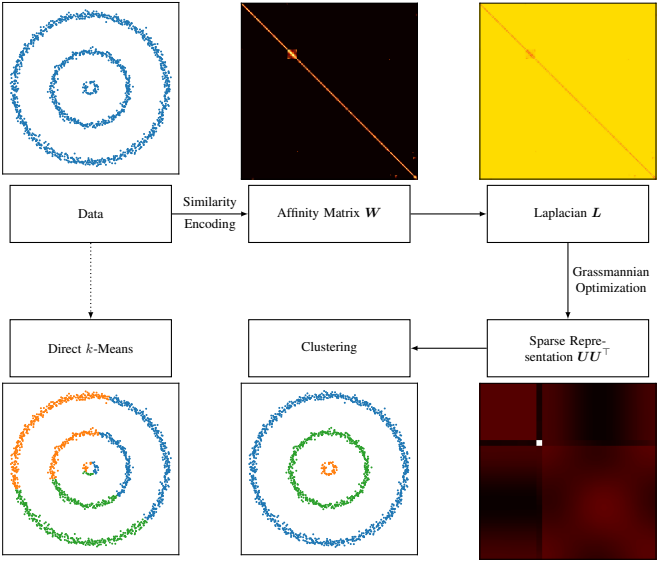}
    }
       \caption{Operations and example of spectral clustering.
       The three-ring data where each cluster has a ring shape, affinity matrix $\W$, unnormalized Laplacian $\L$
       (in logarithmic scale), sparse representation $\U\tp{\U}$
       and the final result. The three diagonal blocks in the $\W$, $\L$, and $\U\tp{\U}$ heat map
       (bottom right) correspond to three clusters.
       Noted direct clustering using $k$-means is ineffective.
}
        \label{fig:ssc_proc}
    \end{figure}

    In this subsection, we will examine how the classic topic of high-dimensional
    data clustering can be cast as a Grassmannian optimization problem.
    \textbf{Spectral clustering} is a  popular technique for dimensionality
    reduction in clustering high-dimensional data, which has been adopted in many
    signal processing applications for its simplicity and performance
    \cite{Chung:1997:SGT,Ng:2002:SCAA,Luxburg:2007:TSC}. The technique derives
    its name from the operation on the spectrum (i.e., the eigenvalues) of the
    Laplacian matrix of the affinity matrix specifying pairwise similarity of
    data samples. The operations  of spectral clustering are illustrated in
    \cref{fig:ssc_proc} and elaborated as follows.

    Given a set of data samples
    $\{x_i\}_{i=1}^N$ from $k$ clusters, Step 1 is to construct an affinity
    matrix $\W\in\sR^{N\times N}$ that contains pairwise sample similarity using
    the distance measure $d(\cdot,\cdot)$ in the original data space. For
    example, $\W$ may be constructed using a Gaussian kernel such that its
    $(i,j)$-th element $\W_{ij} =
    \exp\left\{-\frac{1}{2\sigma^2} d^2(x_i,x_j)\right\}$. The construction of such a matrix, called \emph{similarity encoding},  provides a convenient overview of the similarity heat map for the
    considered dataset as illustrated in \cref{fig:ssc_proc} and \cref{ex:ssc}
    and thereby facilitates the subsequent data clustering.

    As Step 2, the
    similarity ``heat-map'' can be enhanced by constructing the so-called
    Laplacian matrix of $\W$ via $\L = \D - \W$ for unnormalized Laplacian and
    $\L = \I - \D^{-1/2} \W
    \D^{-1/2}$ for normalized one, where $\D \in \sR^{N\times N}$ is a diagonal
    matrix with diagonal entries being the sum of each row in $\W$
    \cite{Shi:2000:NCIS}. The  Laplacian matrix $\L$ is a notation  from graph theory and widely used in computer-vision tasks such as blob and edge detection. Roughly speaking, it captures
    the difference between a specific data sample and the average value of those which are close to it.
    Compared with $\W$, a local measure of similarity,
    the Laplacian $\L$ takes the global structure of the dataset into consideration and thus is expected to achieve a better clustering   performance.

    Step 3 is to eigen-decompose $\L$ and extract data's
    low-dimension representations for subsequent clustering. The $k$
    eigenvectors of $\L$ corresponding to the $k$ smallest  eigenvalues, denoted
    by $\U\in\sR^{N\times k}$, can be viewed as a low-dimensional representation
    that provides the key structural information of the dataset. This allows
    each data sample to be represented by a $k$-dimensional row vector of $\U$.
    It is expected that in this extracted low-dimensional feature space. The data
    clustering structure can be better revealed, thus leading to both
    performance boost and complexity reduction for clustering in the
    subspace (using a standard algorithm such as $k$-means) compared with that
    in the original high-dimensional data space. A simple example is given in
    \cref{fig:ssc_proc} where clustering in the feature space is effective while
    that in the original data space is not.

    \begin{figure}[tb]
	\centering
    \def\svgwidth{0.8\linewidth}
	\input{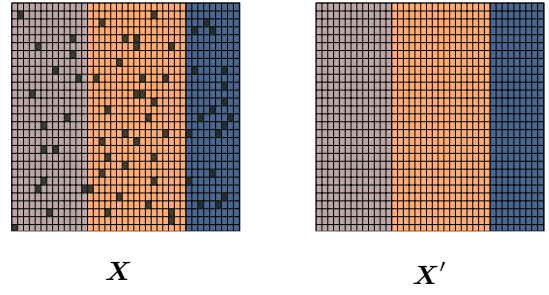}
	\caption{Illustration of low-rank matrix completion.}
	\label{fig:matcomp}
    \end{figure}

    Building on the above basic technique, \textbf{sparse spectral clustering}
    attempts to improve the accuracy in representation by exploiting the
    sparse structure in the affinity matrix $\W$
    \cite{Lu:2016:CSSC}. In the ideal case where a dataset is comprised of well separated clusters,
    data similarity should be local, i.e., data
    from the same cluster are similar to each other but dissimilar to those from
    other clusters. Consequently, the affinity matrix $\W$ should be block diagonal
    (with each block correspond to one cluster) and thus exhibits a sparse
    structure \cite{Jordan:2006:LSC}. It is further argued in \cite{Lu:2016:CSSC}
    that the low-dimensional representation $\U\tp{\U}$ defined earlier should be
    also sparse. Specifically, each row of $\U$ may be a ``one-shot encoded"
    representation of the corresponding data point with only a ``1" at the location
    whose index identifies the corresponding cluster and ``0"s at other locations.
    It follows that samples with identical encoded representations belong to the
    same cluster. Furthermore,
$\U$ is of rank $k$ and also sparse since there are many zeros entries that
    do not contribute to its rank. Given the sparsity, sparse spectral
    clustering can be formulated as an optimization problem for minimizing the
    discrepancy between the Laplacian matrix $\L$ of $\W$ and its
    low-dimensional representation $\U\tp{\U}$ as measured by their inner-product
    plus a penalty term $\beta \left \lVert \U\tp{\U}\right\rVert_0$ that counts the
    number of nonzero elements and thus imposes a sparsity constraint. Since
    optimization over $L_0$ norm is known to be NP-hard, a common relaxation is to
    use the $L_1$ norm instead \cite{Lu:2016:CSSC}. This yields the following
    problem of sparse spectral clustering:
    \begin{equation} \label{eq:ssc}
    \begin{alignedat}{2}
        \underset{\U}{\min} \quad & f \left( \U \right) = \iprod{\U \tp{\U}}{\L}
        +\beta \left \lVert \U \tp{\U} \right \rVert_1 \\
        \text{s.t.} \quad & \tp{\U}\U = \I_k.
    \end{alignedat}
    \end{equation}

 Due to the orthogonality constraint on $\U$, we can cast the problem in \eqref{eq:ssc} into a {\bf Grassmannian optimization} problem based on the framework discussed in \Cref{sec:preliminaries}:
    \begin{equation} \label{eq:grssc}
    \begin{alignedat}{2}
        \underset{\U}{\min} \quad & f \left( \U \right) = \iprod{\U \tp{\U}}{\L}
        +\beta \left \lVert \U \tp{\U} \right \rVert_1 \\
        \text{s.t.} \quad & \U \in \grass{n}{k}.
    \end{alignedat}
    \end{equation}
The optimization in \eqref{eq:grssc} attempts to
    find a desired subspace but not a specific basis, making it a suitable
    application of Grassmannian optimization. To verify this fact, one can observe
    that the objective function in \eqref{eq:grssc} is invariant to any right rotation of
    the variable $\U$ since $\U\R\R^T\U^T =  \U\U^T$ with $\R\in \orth{k}$ being a
    $k\times k$ rotation (unitary) matrix. It follows that the problem in
    \eqref{eq:grssc} can be solved using the procedures for Grassmannian
    optimization described in \Cref{sec:preliminaries}. Based on the resultant
    low-dimensional representation
    of the data samples in $\U$, different clusters can be identified. The classical example of three rings
    is shown in \cref{ex:ssc} to illustrate the power of Grassmannian-aided sparse representation learning.
%---------------------------------------------
%\newpage
\FloatBox*[\linewidth]{
%\small
\vspace{-6mm}
    \begin{example}[Sparse Spectral Clustering of Three-Ring Dataset] \label{ex:ssc} \normalfont
	We consider an three-ring dataset in $\sR^2$ with $N = 1550$ data
	samples forming $k=3$ ring clusters. The clustering problem in
	\eqref{eq:grssc} is solved using the method of steepest
	descent on the Grassmannian with the penalty $\beta=0.01$ and varying $\sigma\in
	\{0.1,1,1.6,3,5\}$, the bandwidth in the Gaussian kernel used for constructing $\W$.
	The rows in \Cref{fig:ex:ssc} are the heat map of $\U\tp{\U}$,
	the embedding of $\U$ in $\sR^3$, and the clustering results and the columns
	different values of $\sigma$. The structural information can be infered from the
	heat maps. For example, starting at $\sigma=1.6$, one can observe three diagonal blocks
	in the heat map, indicating the existence of three clusters. The parameter
	$\sigma$ has a significant effect on the performance, governing
	how the Grassmannian optimization can disentangle data in the low dimensional
	latent space determined by $\U$. For small $\sigma$, the affinities between
	data samples are reduced; data samples are pushed away from each
	other and appear isolated. Increasing $\sigma$
	amplifies similarity and pulls data samples together, resulting in
	parallel circles on the sphere in the latent space. Undesirable clustering
	results arises from either too small or too large $\sigma$ is. With a
	properly chosen $\sigma$ ($1.6$ in this case), all three clusters are
	concentrated in separate small regions in the latent space generated by $\U$
	(second row in \Cref{fig:ex:ssc}). Consequently, data rings are desirably
	clustered as marked in different colours (bottom row in \Cref{fig:ex:ssc}). Note that  in practice,  the optimal value of $\sigma$ can be found by line search.
%\vspace{-3mm}
\begin{center}
	\begin{minipage}[t]{0.05\textwidth}
		\vskip -8cm
		\hspace{10mm} \rotatebox{90}{$\U\tp{\U}$}
		\vskip 2.2cm
		\rotatebox{90}{$\U$}
		\vskip 2.2cm
		\rotatebox{90}{Clustering}
	\end{minipage}%
	\begin{minipage}[t]{0.85\textwidth}
	\includegraphics[width=\textwidth]{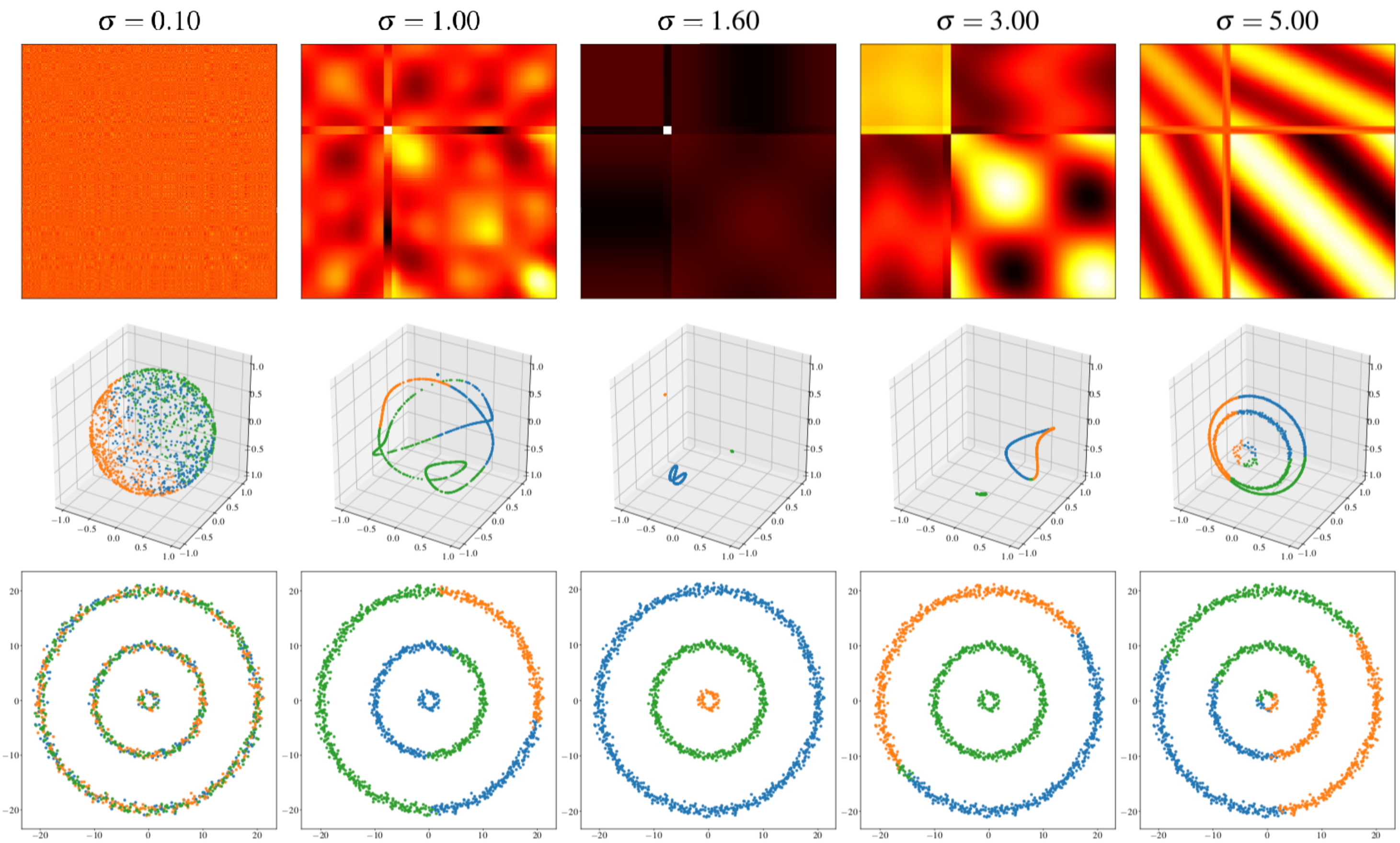}
    \end{minipage}
\end{center}
    %\vspace{-6mm}
    \caption{{Example of sparse spectral clustering via Grassmann optimization.}} \label{fig:ex:ssc}
    %\caption{Illustration of the Example. \textbf{Left: } The objective of the original formulation \eqref{eq:mcp}
	%using Frobenius norm. \textbf{Right: } The objective alternatively formulated using chordal distance.}
	\vspace{-4mm}
    \end{example}
}
\vspace{-2mm}
%---------------------------------------------

\begin{figure*}[t]
\begin{center}

       \begin{subfigure}[t]{0.4\textwidth} \vspace*{2mm}
            \centering
        \resizebox{0.7\textwidth}{!}{
            \includegraphics{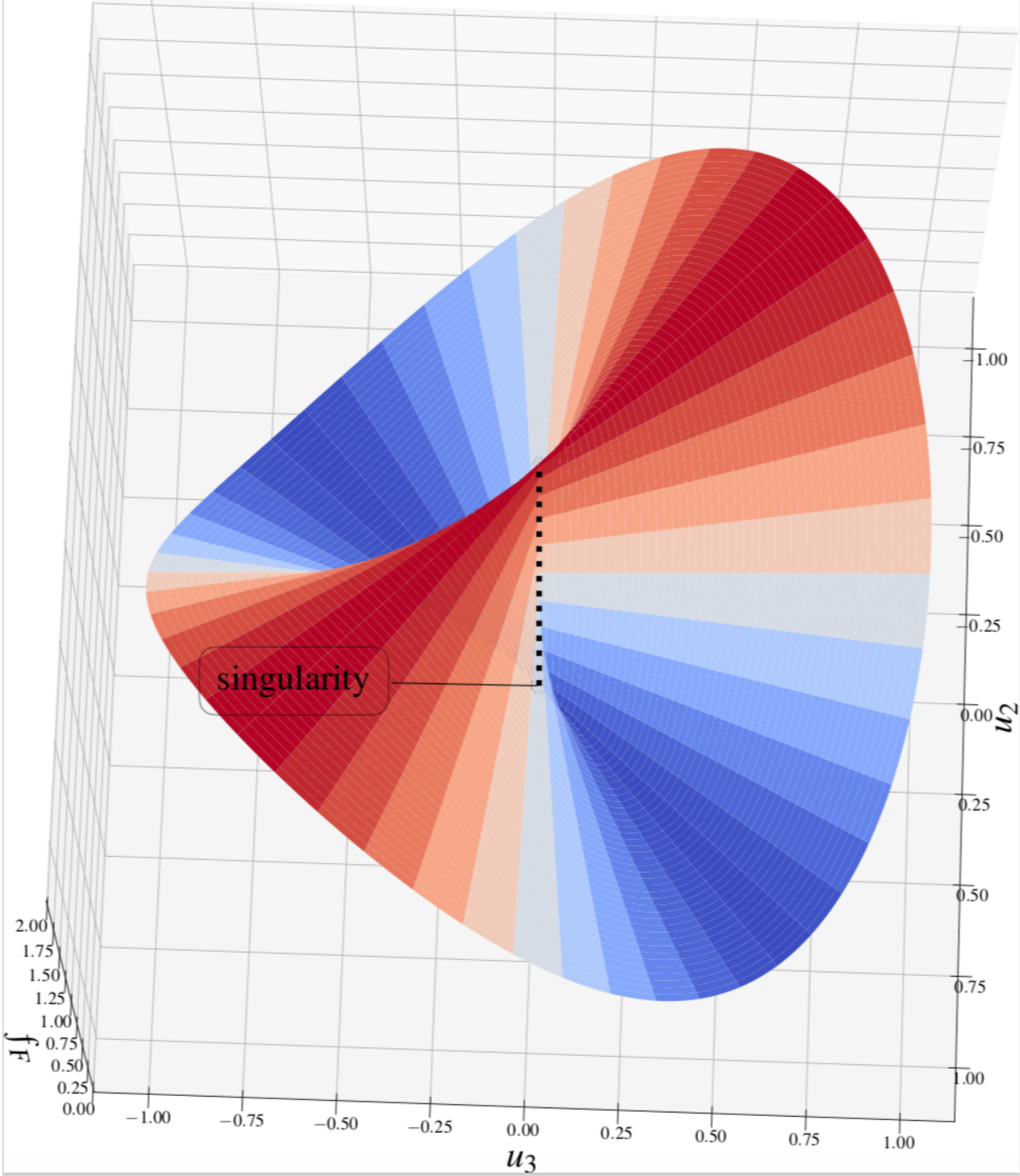}
        }
        \end{subfigure}%
        \begin{subfigure}[t]{0.4\textwidth} \vspace*{2mm}
            \centering
        \resizebox{0.7\textwidth}{!}{%
        \includegraphics{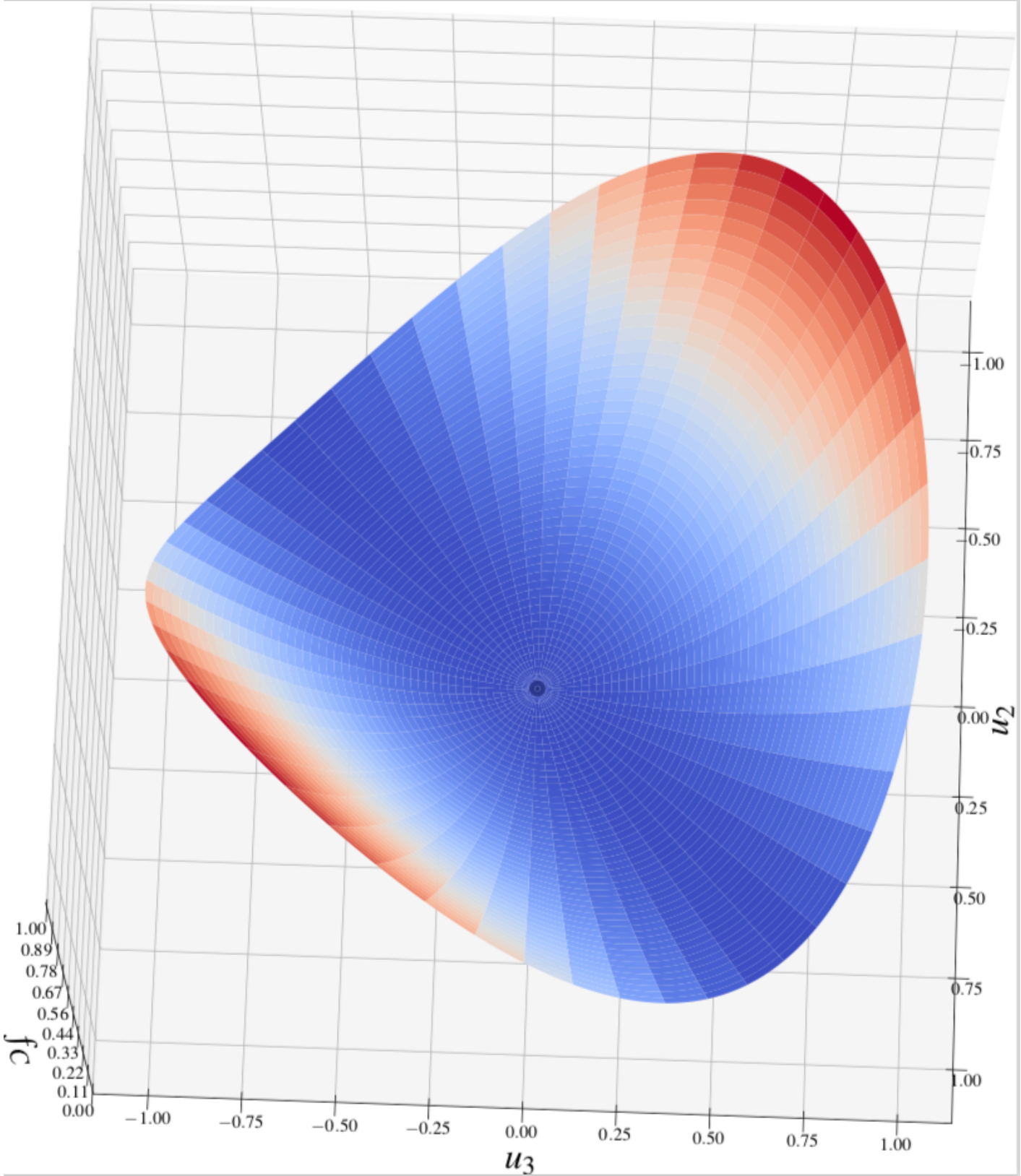}
        }
        \end{subfigure}
        \caption{Objective function profiles for (left) Frobenius-distance formulation
       and (right) projection-distance formulation.} \label{mc_illustration}

\label{Fig:Singularity}
\end{center}
\end{figure*}

\subsection{Low-Rank Matrix Completion}
\label{sec:trad:lrr}

    Another classic problem in low-rank representation learning is {\bf low-rank matrix completion},
    namely filling in the missing entries of a given matrix. Though the general
    problem is ill-posed, it is tractable under the assumption that the matrix to be filled is low-rank.
    We give an illustration of the problem in Fig. \ref{fig:matcomp}. The incomplete
    matrix $\X$ has a rank of $3$ as characterized by three one-rank blocks in
    different colours. The black rectangles correspond to the missing entries, which
    can be filled in based on consistency with the observed entries. Thereby, the
    complete matrix $\X'$ can be reconstructed from the partially observed matrix
    $\X$.

    Mathematically, the problem of low-rank matrix completion can be formulated as follows.
    Consider a matrix $\X \in \sR^{n\times k}$ with $k \le n$ and a
    constraint on its rank $r \le \min \{n, k\}$. Let us define an indicator
    matrix $\Omega_{\X}$ for observed entries such that $\Omega_{ij} = 1$ if
    $\X_{ij}$ is observed and $0$ otherwise. The aim is to find a  low-rank
    reconstruction of $\X$ that is consistent with the observed entries in
    $\X$. More formally, let $\mathcal{P}_{\Omega}$ denote a mapping that maps all unobserved entries to $0$ and keeps the observed entries:
    \begin{equation} \label{def:matcomp}
	\mathcal{P}_{\Omega}: \X \mapsto \X_{\Omega} =
	\begin{cases}
	    \X_{ij}, & \Omega_{ij} = 1, \\
	    0, & \text{otherwise}. \\
	\end{cases}
	\end{equation}
    The matrix-completion problem seeks a matrix $\X' \in \sR^{n\times k}$,
    such that $\rank{\X'} = r$ and $\mathcal{P}_{\Omega}(\X') = \mathcal{P}_{\Omega}(\X)
    = \X_{\Omega}$.
    This problem is usually reformulated as a \emph{subspace identification}
    problem that searches for the column space of the desired matrix $\X$,
    denoted as $\U$, that is consistent with that of the observed matrix
    $\X_{\Omega}$ by finding the minimum  Frobenius-norm based distance  $f_F (\U, \X_{\Omega})$:
    \begin{equation} \label{eq:mcp}
    \begin{alignedat}{2}
    \text{solve} \quad & f_F (\U, \X_{\Omega}) = \min_{\W\in\sR^{k\times r}} \left
	\lVert \X_{\Omega} - \mathcal{P}_{\Omega} \left( \U \tp{\W} \right) \right\rVert_F^2= 0 , \\
	\text{s.t.} \quad & \tp{\U} \U = \I_n,\\ %\in \grass{n}{r}. \\
    \end{alignedat}
    \end{equation}
    where the right-multiplication of $\W$ on $\U$ amounts to a proper
    transformation to align $\X$ with the coordinates of $\X_{\Omega}$ within the
    subspace $\U$. It is found in \cite{Dai:2012:GLRC} that the use of Frobenius
    norm in (\ref{eq:mcp}) introduces singularities that may cause difficulty in
    applying the gradient based methods to search for the global minimum
    \cite{Dai:2012:GLRC}.

    This issue can be fixed by recasting the
    the problem in (\ref{eq:mcp}) as a {\bf Grassmannian optimization problem} that
    leverages the geometry of the original formulation, namely that the optimization
    variable $\U$ is more properly modelled as an element on some Grassmann manifold
    rather than an ordinary orthonormal matrix. The reason is that the minimization
    over $\W$ makes the objective function $f_F(\U, \X_{\Omega})$ invariant to rotation of $\U$,
    which makes the problem in \eqref{eq:mcp} defined on the Grassmannian rather
    than that in \eqref{eq:ssc} on the Stiefel manifold. For the reformulation, a required step is to design a proper measure of
    the subspace-distance between the low-rank representation $\U$ and the
    partially observed matrix $\X_{\Omega}$, denoted as $f_S (\U, \X_{\Omega})$,
    such that the measure is everywhere continuous without any singularity. This
    facilitates the solving of the following reformulated Grassmannian-optimization
    problem
    \begin{equation}
    \begin{alignedat}{2}
	\text{min} \quad & f_S (\U, \X_{\Omega}), \\
	\text{s.t.} \quad & \U \in \grass{n}{k}\\
    \end{alignedat}
    \end{equation}
    using the gradient based method discussed in Section \ref{sec:pre:op}.

    One such design is proposed in \cite{Dai:2012:GLRC} and briefly described as follows.
     Let $\x_i$ denote the $i$-th of a total of $n$ rows of the partially observed matrix
    $\X_{\Omega}$ where the missing entries are filled with zeros.
    A matrix $\B_i$ is then constructed by cascading the column vector $\x_i$
    and a set of basis vectors $\{\e_j\}$, each of which has a ``$1$" at the
    $j$-th entry and $0$s elsewhere:
    \begin{equation}
	\B_i = [\x_i, \e_{j}, \e_{j'}, \cdots]
    \end{equation}
    where $\{\e_j\}$ are chosen such that the rows of $\B_i$ corresponding to the
    missing entries in $\x_i$ each has a ``$1$" entry and zeros elsewhere. Then an
    objective function enabling Grassmannian optimization is designed as follows:
    \begin{align}{\label{chordal formulation}}
	 f_P(\U; \X_{\Omega}) =  \sum_{i=1}^n d_P(\B_i, \X_{\Omega}),
    \end{align}
    where $d_P(\A, \B) = 1 - \lambda_{\max}(\A^T \B)$ is the projection distance, a
    subspace distance measure, between two matrices $\A$ and $\B$ and
    $\lambda_{\max}(\cdot)$ gives the maximum singular value of its argument. As
    illustrated in \Cref{Fig:Singularity}, the new objective function $f_P(\U;
    \X_{\Omega})$ based on subspace distance has a smooth profile while that of the
    original one based on F-norm has singularities causing difficulty in conducting
    gradient descent algorithm.

    \begin{figure*}
	\centering
	\def\svgwidth{0.7\textwidth}
	\input{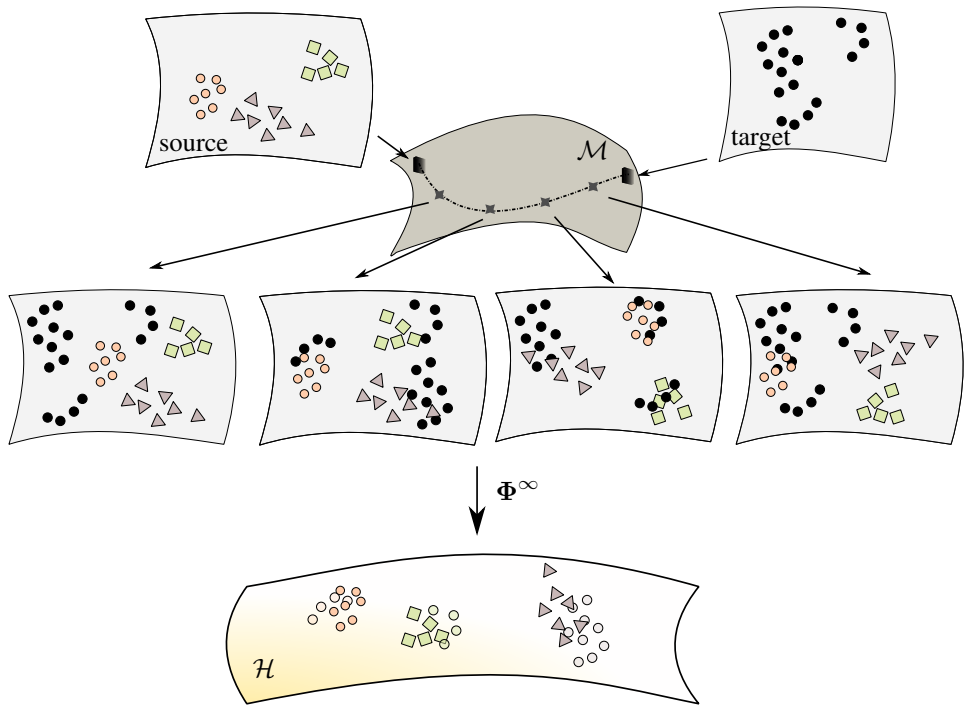}
	\caption{Geodesic-sampling based transfer learning.}
	\label{fig:gfk}
    \end{figure*}

\section{Deep Grassmannian Learning}
\label{sec:deep}

    The revival of artificial neural networks and deep learning has achieved
    unprecedented success in a wide spectrum of applications including computer
    vision, natural language processing and artificial intelligence
    \cite{Lecun:2015:DL}. Recently, researchers started to develop geometric
    techniques such as geodesic convolution \cite{Masci:2015:GC} and matrix
    back-propagation \cite{Ionescu:2015:MBP} for fully unleashing the potential of
    deep neural networks in solving problems having embedded geometric structures
    \cite{Bronstein:2017:GEODL}. In the same vein, one may ask how Grassmann
    manifolds can be incorporated into deep learning to streamline its operation or
    improve its performance in certain applications, leading to deep Grassmannian
    learning. In this section, two specific areas of deep Grassmannian learning are
    introduced, namely visual domain adaptation (or transfer learning) in
    \Cref{sec:deep:tl} and the construction of
    deep neural networks for Grassmannian inputs or outputs \Cref{sec:deep:net}.

\subsection{Transfer Learning}
    \label{sec:deep:tl}
    Transfer learning refers to the task of generalizing the knowledge of a
    model learned in one domain (source) to another domain (target) e.g., from
    handwriting recognition to street-sign recognition, or from natural language
    processing in one language to that in another. Currently, the most prevalent
    methods lie in one of three categories: \cite{Patel:2015:DASURVEY}
    \begin{enumerate}
    \item (Dimensionality Reduction) The methods are based on the
	assumption that datasets from different domain share similar
	representations in a certain latent feature space that can be recovered
	using a  dimensionality reduction technique;
    \item (Data Augmentation) The principle of these methods (such as
	geodesic-flow methods discussed in the sequel) is to intelligently mix
	the datasets from the source and target domains such that the latent
	feature space mentioned earlier becomes not only explicit but even
	dominant, allowing its extraction.
    \item (Deep Learning) The methods involves the deployment of deep
	neural networks in transfer learning and may not be mutually exclusive
	with previous two types.
\end{enumerate}

    This subsection focuses on a key topic of transfer learning, namely {\bf visual
    domain adaptation}, and its relevance to Grassmannian learning. One example is
    using the model for recognizing objects in the images on Amazon to recognize
    objects in the wild. To be precise, a typical task of visual domain adaptation
    is defined as follows. Given $N$ labeled data samples in the source domain
    encompassing $C$ classes with $N_c$ observations in each class. Therefore, the
    dataset can be represented by $\left\{\s_{c,i}: c=1,2,\ldots, C, i=1,2,\ldots,
    N_c \right\}$. Then (unsupervised) transfer learning aims at predicting the
    class labels of the observations in the target domain where data samples are
    unlabeled. In the remainder of this subsection, we will discuss an approach in
    Grassmannian deep learning for performing this task.

\subsubsection{Geodesic-Flow Methods}

    The geodesic-flow methods rely on the assumption that the same class of
    images in two different domains may be modelled as separate points
    (subspaces) on a low-dimensional Grassmann manifold. An intermediate
    subspace on the Grassmann geodesic linking the domains may be viewed as a
    reasonable latent feature space for learning a common representation.
    Unlike many other methods inline with this philosophy
    \cite{Patel:2015:DASURVEY}, the geodesic-flow methods model the source and
    target domains as points on the Grassmann manifold. We consider two methods
    in the discussion: \emph{sample geodesic flow} (SGF) \cite{Gopalan:2011:GGD}
    and GFK \cite{Gong:2012:GFK}. As
    illustrated in Fig.~\ref{fig:gfk}, these methods explore intermediate latent
    spaces by sampling the geodesic joining the two domains. Based on the same
    principle, SGF and GFK differs in how the latent space is generated from the
    geodesic.

    First, consider the SGF method that aims at seeking discrete and finite
    samples of the geodesic. The idea is illustrated in the upper half of Fig.
    \ref{fig:gfk}. By traversing along the geodesic, the distributions of
    features (or latent representations) from two domains may be most similar at
    some location, where the classes in the source domain can be identified
    leveraging the classifier from the source domain.
    Mathematically, the geodesic from the source domain $\X_s$ to the target domain
    $\X_t$, denoted as $\bPhi$ ,  can be constructed using the CS decomposition
    \eqref{eq:csd} introduced in \Cref{sec:pre:geo}. Let the tangent at $\X_s$ on
    the Grassmannian be denoted as $\bDelta$. The geodesic can be written as
    \begin{equation} \label{eq:gfk:geodeisc}
    \begin{aligned}
        \bPhi(t) = \X_s \U_1 \bGamma(t) \tp{\V} - \bDelta \U_2 \bSigma(t) \tp{\V}.
    \end{aligned}
    \end{equation}
    % where $\bGamma(t) = \diag{\cos \theta_i t}$, $\bSigma(t) = \diag{\sin \theta_i}$, and
    % $\theta_i$'s are principal angles.
    Given the geodesic, SGF samples a finite number of points
    on the geodesic, $\{\bPhi(t_n)\}$, and uses the one that performs the best
    (based on  e.g., a line search) to transfer the
    source classifier to label the target-domain data samples.

   \FloatBox*[\linewidth]{
      \vspace{-6mm}
	\begin{experiment}[Transfer Learning Using SGF/GFK] \label{ex:tl:surf} \normalfont
    The source and target domains are
     \texttt{dslr} (red framed-boxes) and \texttt{webcam} (green framed-boxes)
     datasets from the Office dataset.
     Images in the two categories "bike" and "mug" are used in the
     experiment.
    The visualization relies on the classic \emph{$t$-distributed stochastic
    embedding} ($t$-SNE) algorithm for projecting (image) data points onto the paper
    ($\sR^2$). As the input for visualization, the image features are extracted
    using the well-known SURF feature extractor. The experimental results are shown
    in the figures below. (\textbf{Upper Left}) The original SURF feature
    distribution of the images show that the clusters of "bikes" and "mugs" are not
    differentiable. (\textbf{Upper Right}) By applying GFK, the dominant feature
    space is identified as shown in the subfigure. One can observe that "bikes" and
    "mugs" are well separated into clusters, the clusters in different domains but
    the same category are aligned. The results from SGF at the locations of $t =
    0.4$ and $t = 0.6$ on the geodesic are shown in \textbf{Lower left} and
    \textbf{Lower Right}, respectively. One can observe that the $t = 0.6$ result is
    better than that for $t = 0.4$ and approaches the performance of GFK.
        %/\vspace{-3mm}
    \begin{center}
        \begin{minipage}[t]{0.4\textwidth}
            \centering
        \includegraphics[width=0.75\textwidth]{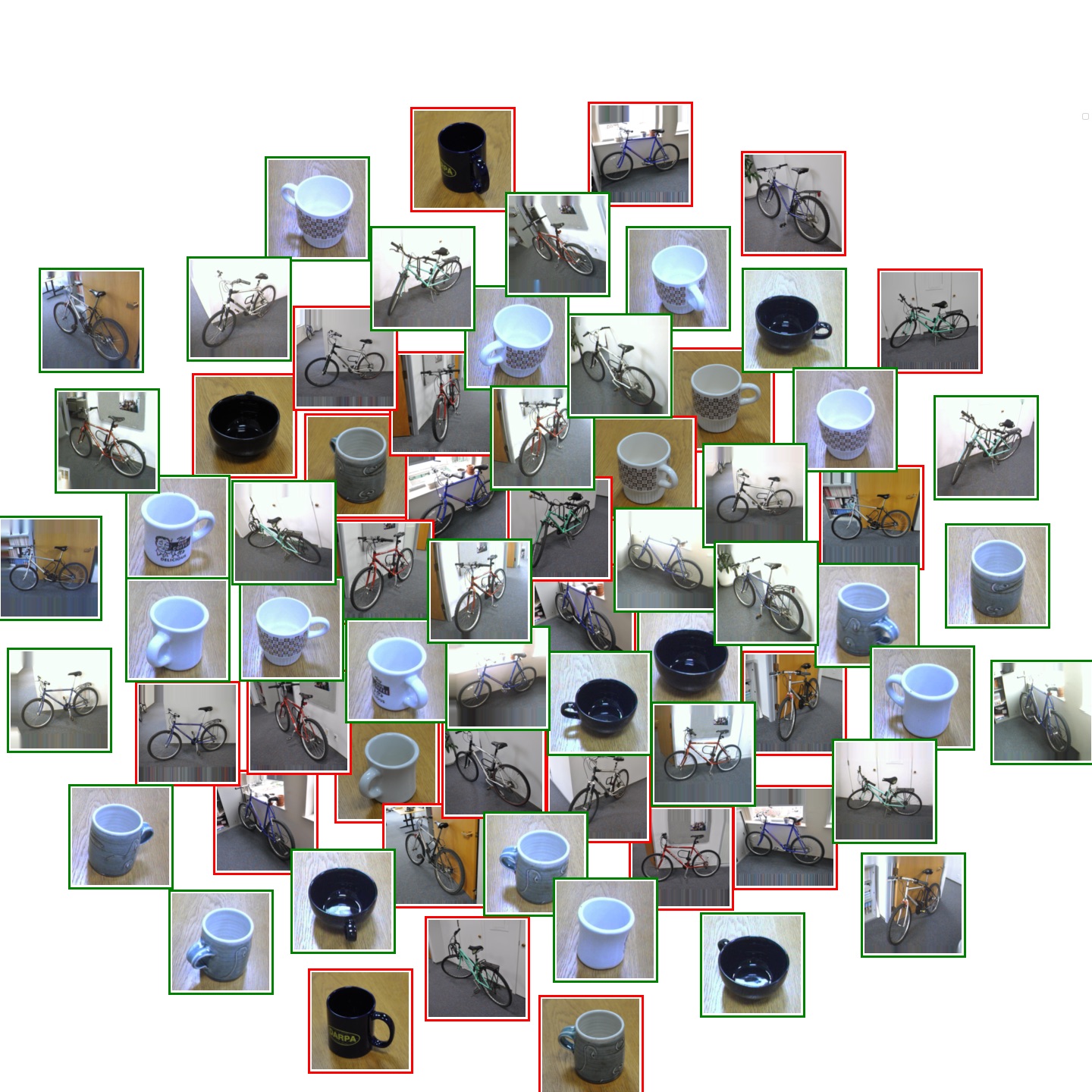}
        \caption*{\small (a) Original (SURF) feature space.}
        \end{minipage}
        %\vrulesep%
        \begin{minipage}[t]{0.4\textwidth}
            \centering
        \includegraphics[width=0.75\textwidth]{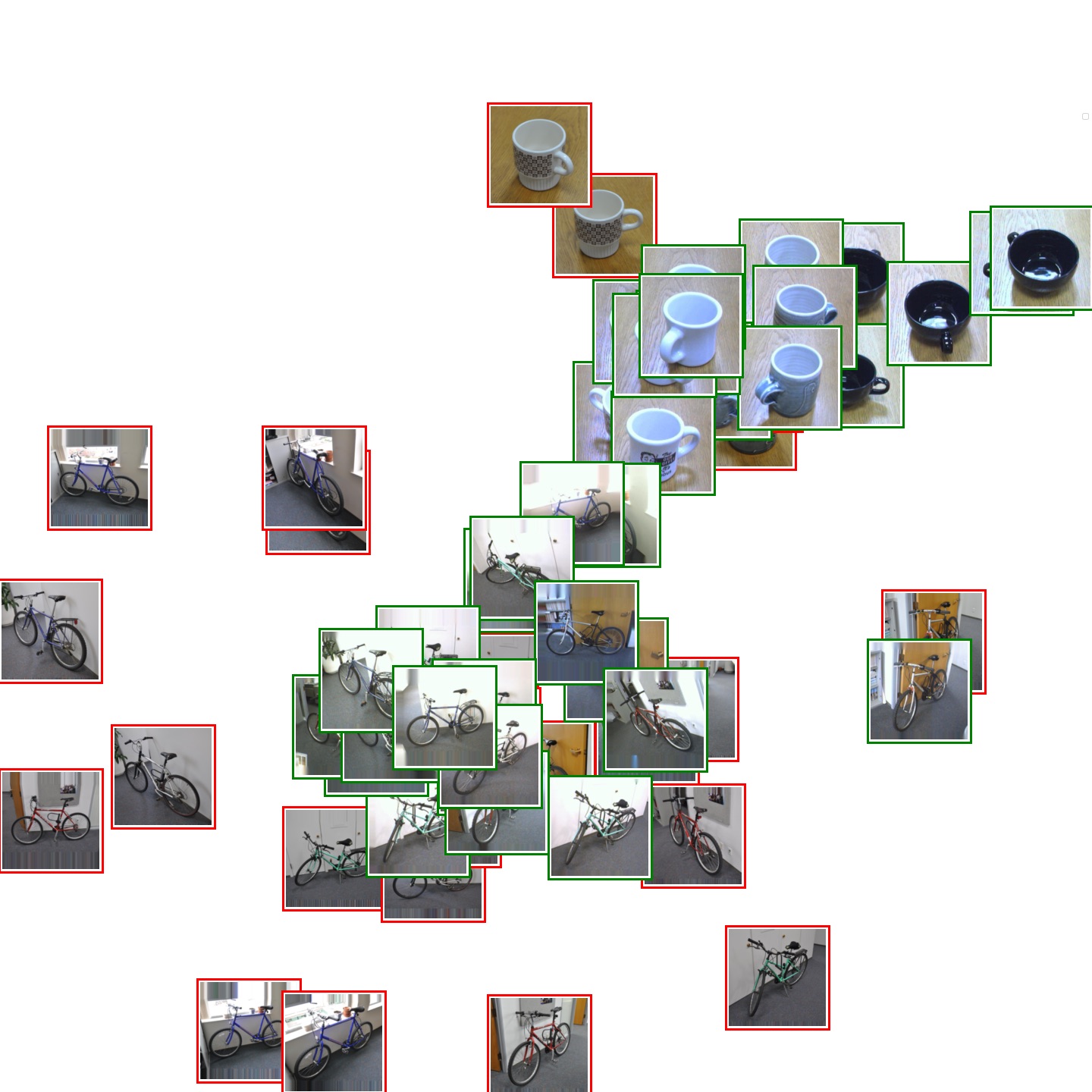}
        \caption*{\small (b) Feature space from GFK.}
        %\label{fig:surf_tsne:gfk}
        \end{minipage}
        %\vspace{-4mm}
        %\hrule
        \begin{minipage}[t]{0.4\textwidth} \vspace*{1mm}
            \centering
        \includegraphics[width=0.75\textwidth]{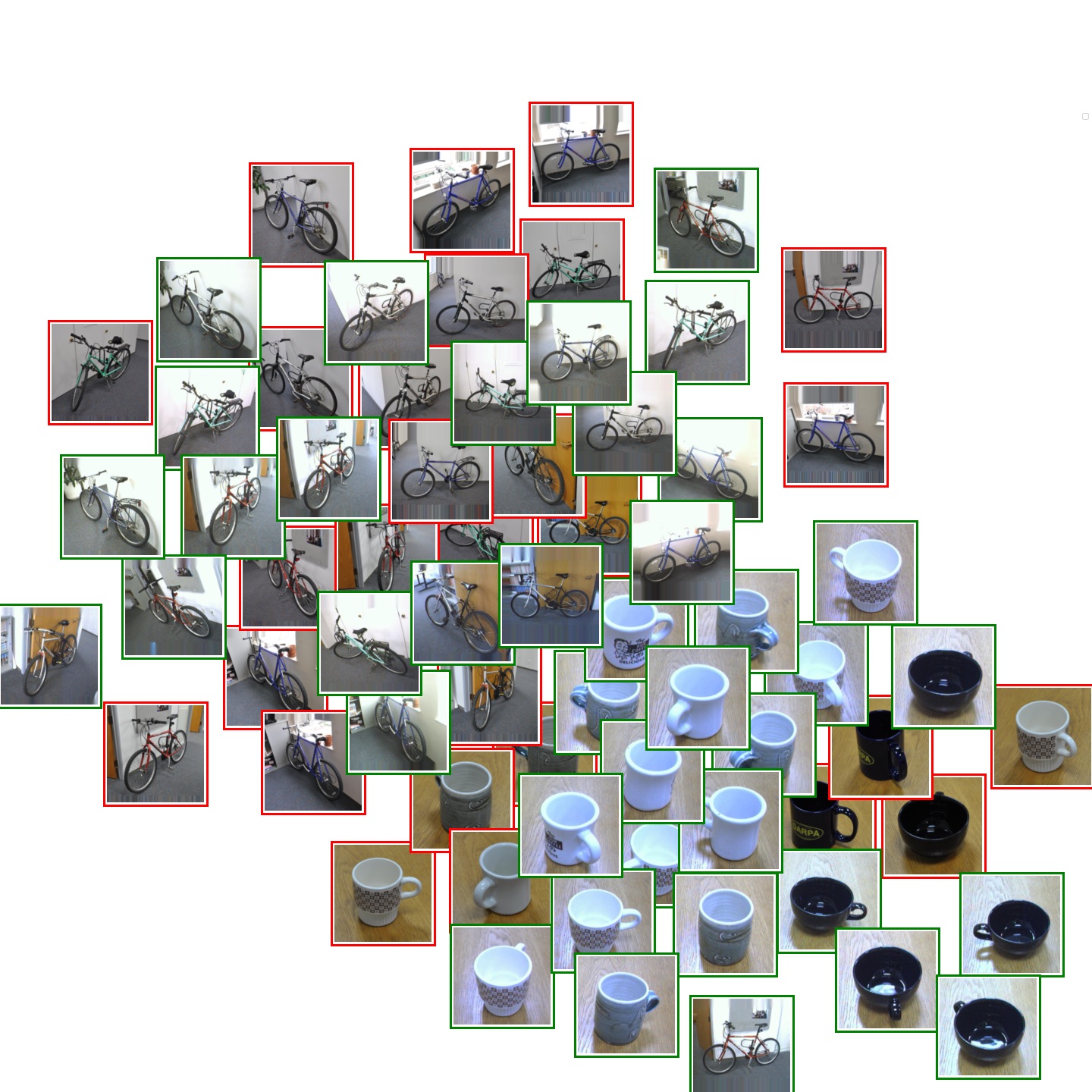}
        \caption*{\small (c) Feature space from SGF with $t=0.4$.}
        %\label{fig:surf_tsne:smp1}
        \end{minipage}
        %\vrulesep%
        \begin{minipage}[t]{0.4\textwidth} \vspace*{1mm}
            \centering
        \includegraphics[width=0.75\textwidth]{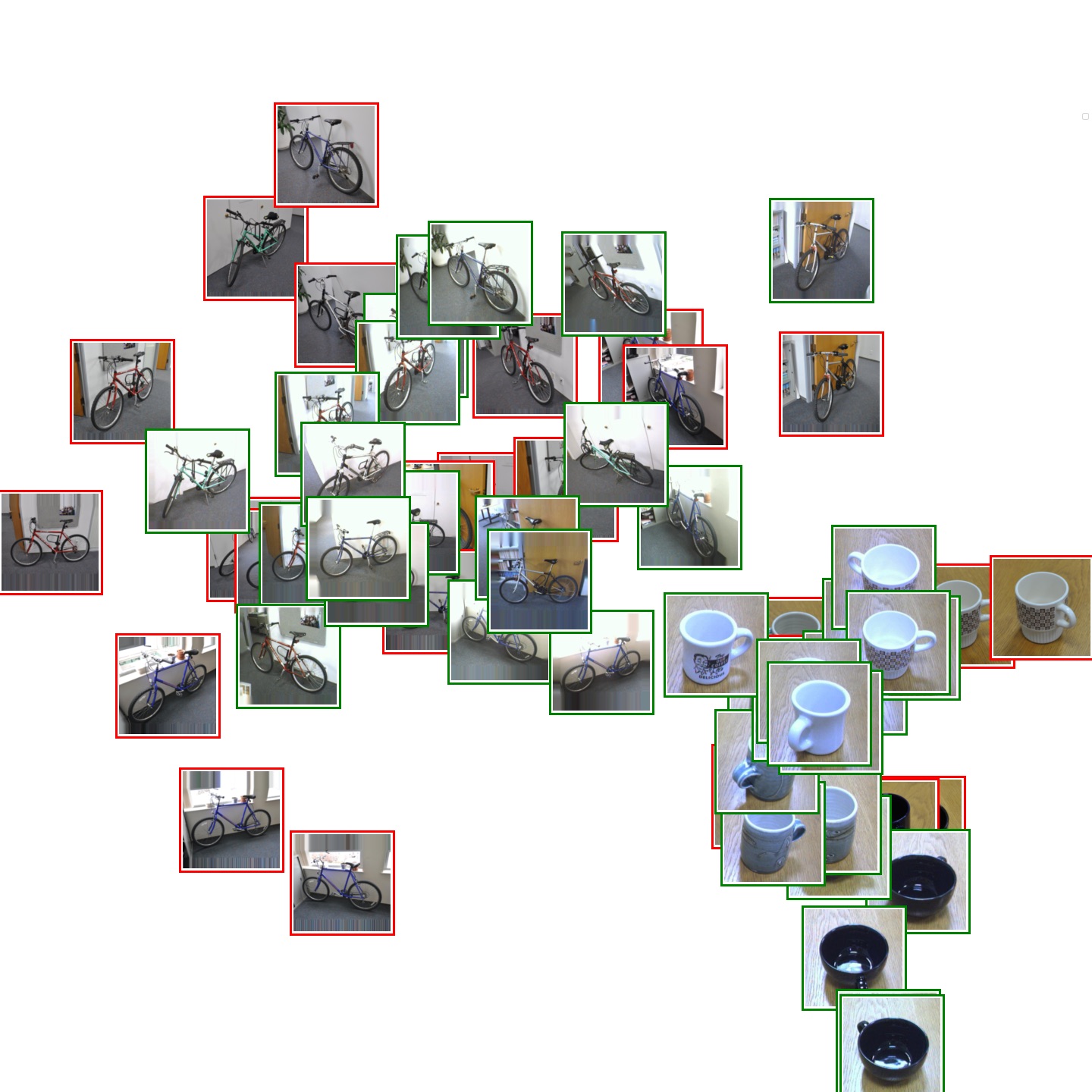}
        \caption*{\small (d) Feature space from SGF with $t=0.6$.}
        %\label{fig:surf_tsne:smp2}
        \end{minipage}
    \end{center}
    \end{experiment}
    }

    Next, consider the GFK method. By exploiting the kernel method, GFK provides an
    elegant way of data augmentation (integrating datasets from source and target
    domains) by making use of all intermediate subspaces across the geodesic.
    The mathematical principle of GFK is to introduce a latent RKHS by integrating
    over the geodesic and thereby constructing a so called \emph{geodesic-flow
    kernel} as illustrated in the lower half of Fig. \ref{fig:gfk}. Specifically,
    the geodesic-flow kernel matrix $\K$ can be computed by
    \begin{equation} \label{eq:gfk:kernel}
	\begin{aligned}
	    \tp{\x_i}\K \x_j = \int_{0}^{1} \tp{\left( \tp{\bPhi(t)} \x_i\right)} \left( \tp{\bPhi(t)} \x_j \right) \dd t.
	\end{aligned}
    \end{equation}
    Essentially, by integrating along the geodesic to construct the flow kernel, all
    significant features that are discriminant in both domains are
    amplified and retained in the infinite-dimensional RKHS. This enables an
    effective knowledge transfer from the source domain to the target domain. As a
    result, GFK outperforms SGF.

    To illustrate the effectiveness of GFK and SGF, an experiment has been carried
    out and the results are presented in \Cref{ex:tl:surf}.

    \subsubsection{Integrating Geodesic-Flow and Deep Learning}
    It is widely known that the quality of features extracted from data impacts
    learning performance. Considering the power of deep neural networks in feature
    extraction, a natural and interesting question to ask is: Is it helpful to
    integrate the preceding geodesic-flow methods with deep neural networks? Indeed,
    a geodesic-flow method can be applied on features extracted using deep neural
    networks to improve the transfer-learning performance. To substantiate this
    point, two experiments have been conducted that show significant performance
    improvements contributed by deep learning. The detailed results are presented in
    \cref{ex:tl:vgg,ex:tl:table}.

    \FloatBox*[\linewidth]{
            \vspace{-6mm}
    \begin{experiment}[SGF/GFK with Deep Feature Extraction] \label{ex:tl:vgg} \normalfont
	Consider the settings in Experiment \ref{ex:tl:surf} but the features of ``mug''
	and ``bike' images are now extracted using a deep neural network instead of
	SURF, a shallow-learning technique. Specifically, the VGG-16 model (a 16-layer
	neural network) \cite{Simonyan:2014:VGG} pretrained on the large-scale image
	database called \emph{ImageNet} \cite{Deng:2009:IMAGENET} is applied to perform
	feature extraction, which generates $4096$-dimensional features for each image.
	Then the features are fed into GFK/SGF transfer learning. The visualization of
	the results are provided in the figures below. (\textbf{Upper left}) Like the
	SURF features in \cref{ex:tl:surf}, the distributions of the original features
	(from the second last fully-connected layer of VGG-16) cannot directly lead to
	the separation of ``bikes" and ``mugs". (\textbf{Upper left}) Nevertheless, the
	application of GFK extracts a feature space where the two categories are
	perfectly separated and the representations in the source and target domains are
	well aligned. One can observe substantial performance gain of deep learning over
	shallow learning in \cref{ex:tl:surf}. Similar conclusions can be drawn for SGF
	with $t = 0.4$ (\textbf{Lower left}) and $t = 0.6$ (\textbf{Lower right}).
	Comparing GFK and SGF, the former significantly outperforms the latter in the
	current context of deep learning.
	%Under the same settings as in \Cref{ex:tl:surf} but the features are now
	%extracted using the VGG-16 model (a 16-layer deep neural network)
	%\cite{Simonyan:2014:VGG} pretrained on \emph{ImageNet}
	%\cite{Deng:2009:IMAGENET} for feature extraction instead of the SURF.  We use
	%the second fully-connected layer (\texttt{FC2}) for this extraction.  The
	%visualization of the results are provided below. (\textbf{Upper left}) Similar
	%to the SURF features in in \cref{ex:tl:surf}, the distributions of the original
	%features may be not immediate clear how ``bikes" and ``mugs" can be separated.
	%(\textbf{Upper left}) Nevertheless, the use of GFK separates the two categories
	%and aligned.  The $t$-SNE plots thus indicates the deep features may bring
	%substantial performance gain over shallow learning in \cref{ex:tl:surf}.
	%Similar conclusions can be drawn for SGF with $t = 0.4$ (\textbf{Lower left})
	%and $t = 0.6$ (\textbf{Lower right}). Comparing GFK and SGF, the former
	%significantly outperforms the latter in the current context of deep learning.
    \begin{center}
        \begin{minipage}[t]{0.4\textwidth}
            \centering
        \includegraphics[width=0.75\textwidth]{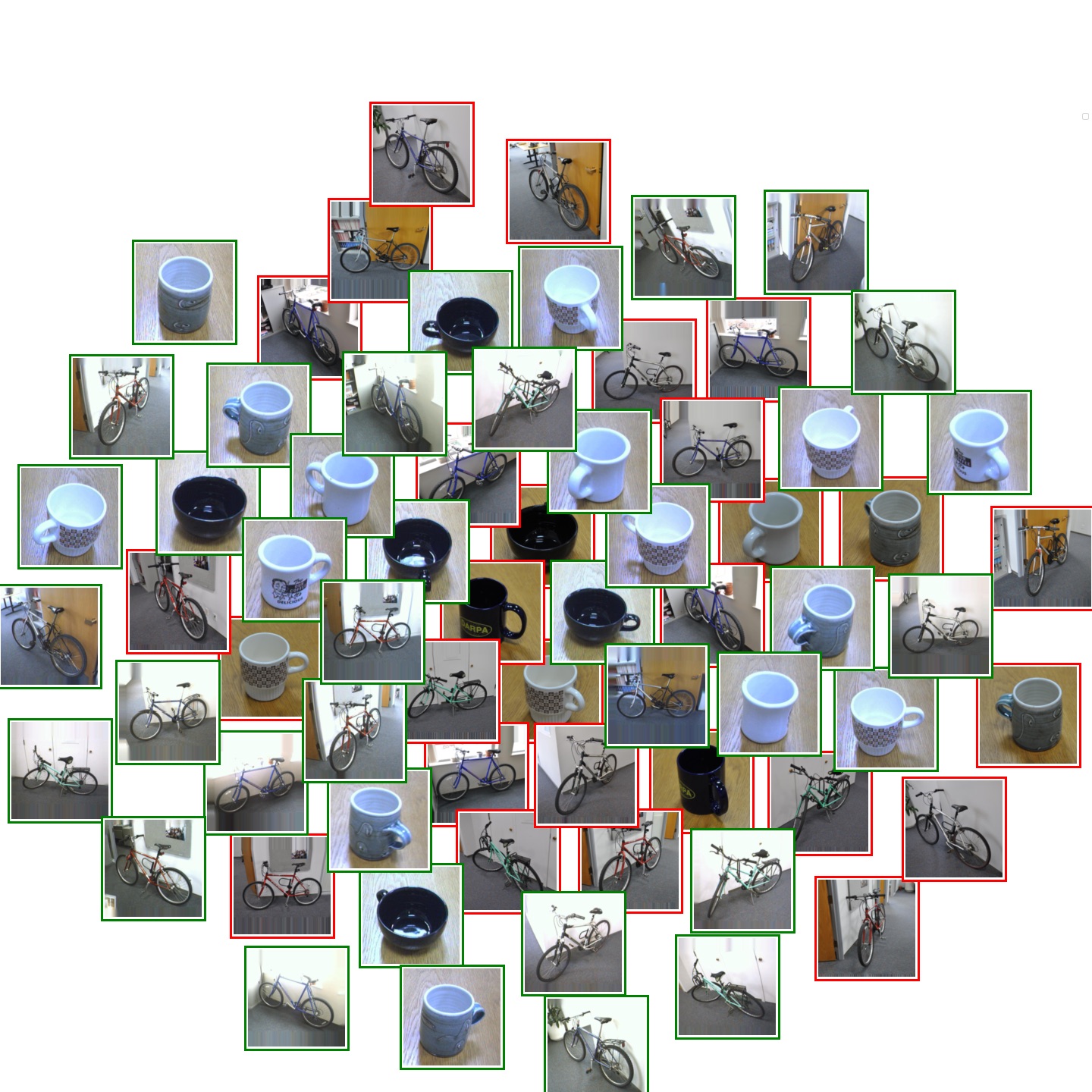}
        \caption*{\small (a) Original VGG feature space.}
        \end{minipage}
        %\vrulesep%
        \begin{minipage}[t]{0.4\textwidth}
            \centering
        \includegraphics[width=0.75\textwidth]{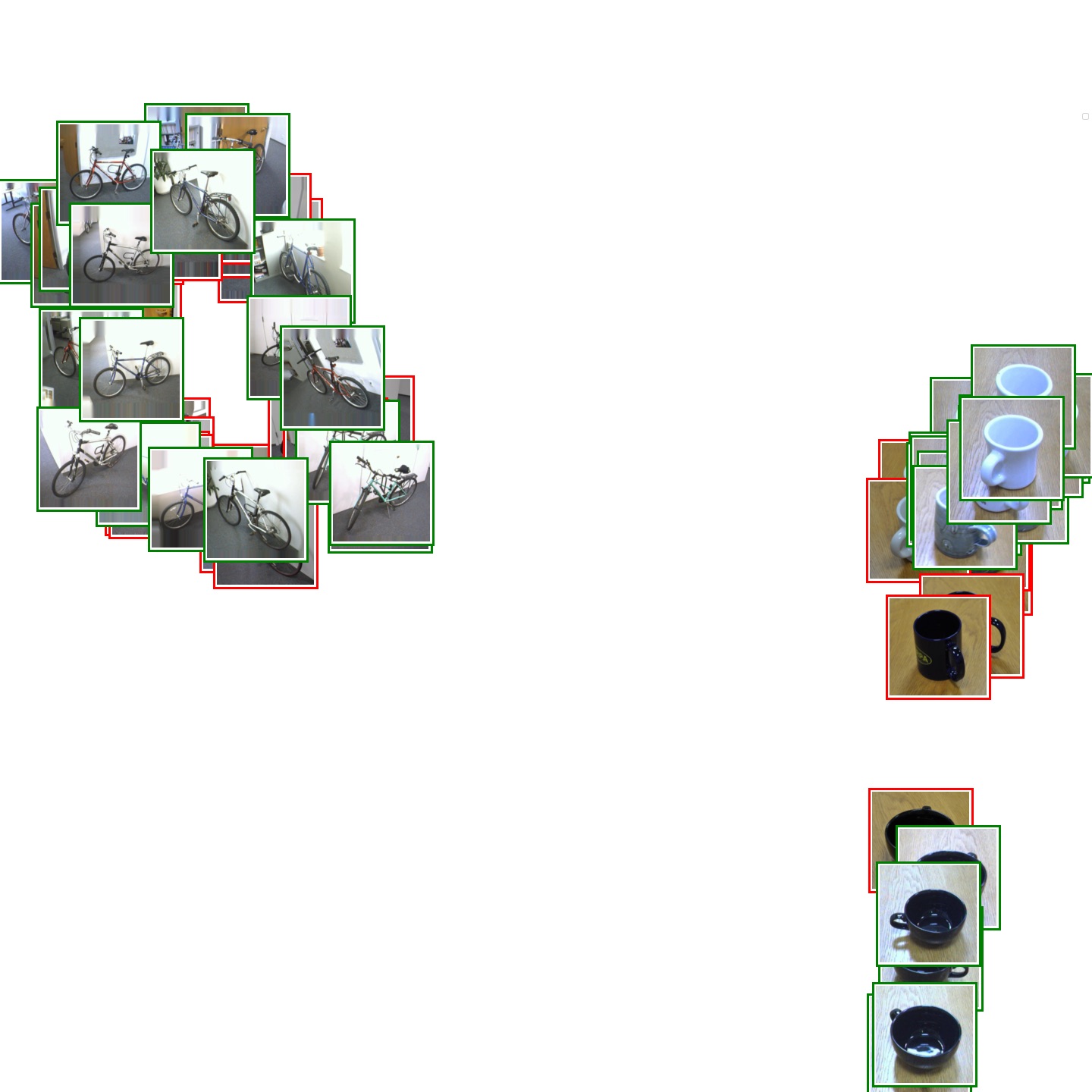}
        %\vspace{-4mm}
        \caption*{\small (b) Feature space from GFK.}
        %\vspace{1mm}
        \end{minipage}
        %\hrule
        \begin{minipage}[t]{0.4\textwidth} \vspace*{1mm}
            \centering
        \includegraphics[width=0.75\textwidth]{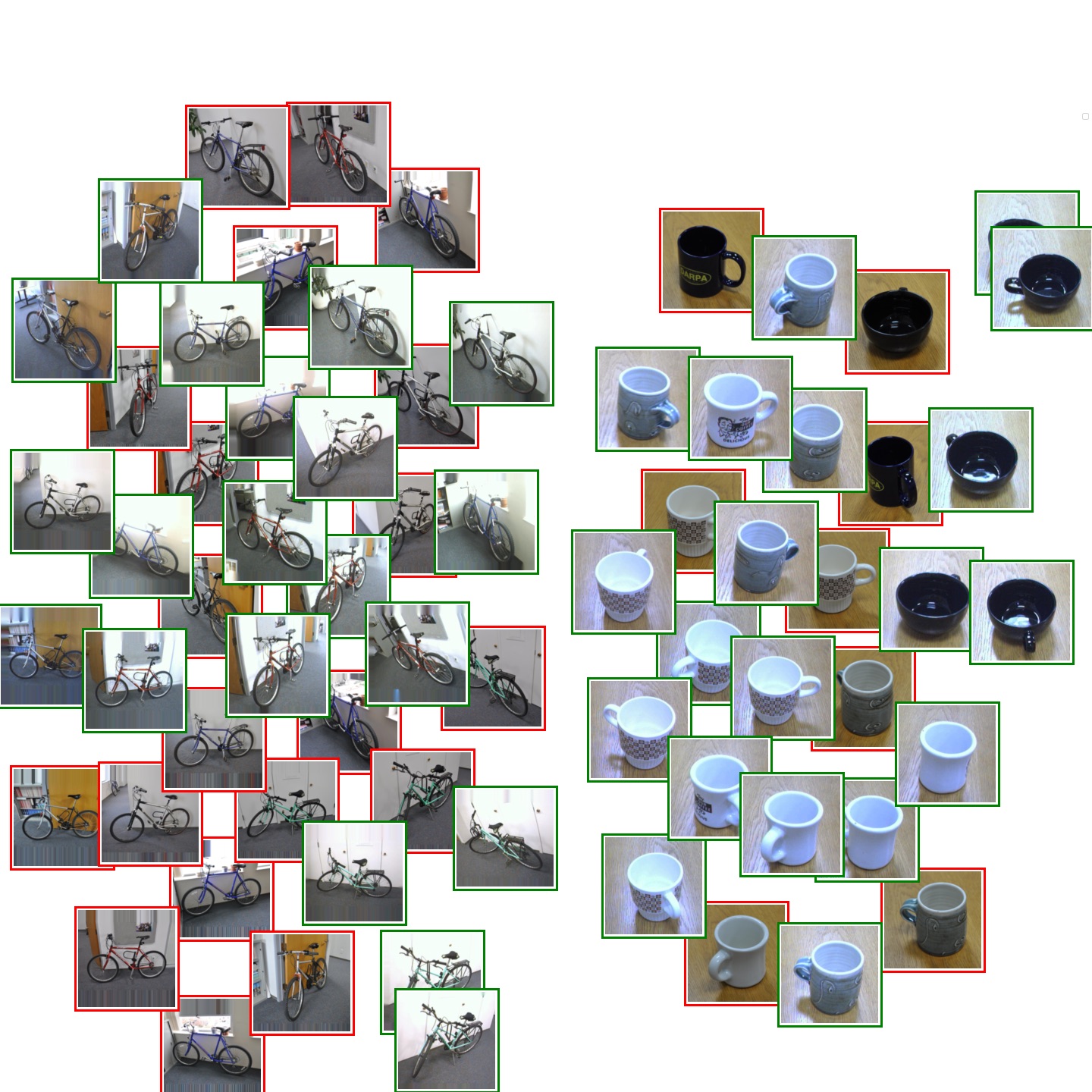}
        %\vspace{-4mm}
        \caption*{\small (c) Feature space from SGF with $t=0.4$.}
        \end{minipage}
        %\vrulesep%
        \begin{minipage}[t]{0.4\textwidth}\vspace*{1mm}
            \centering
        \includegraphics[width=0.75\textwidth]{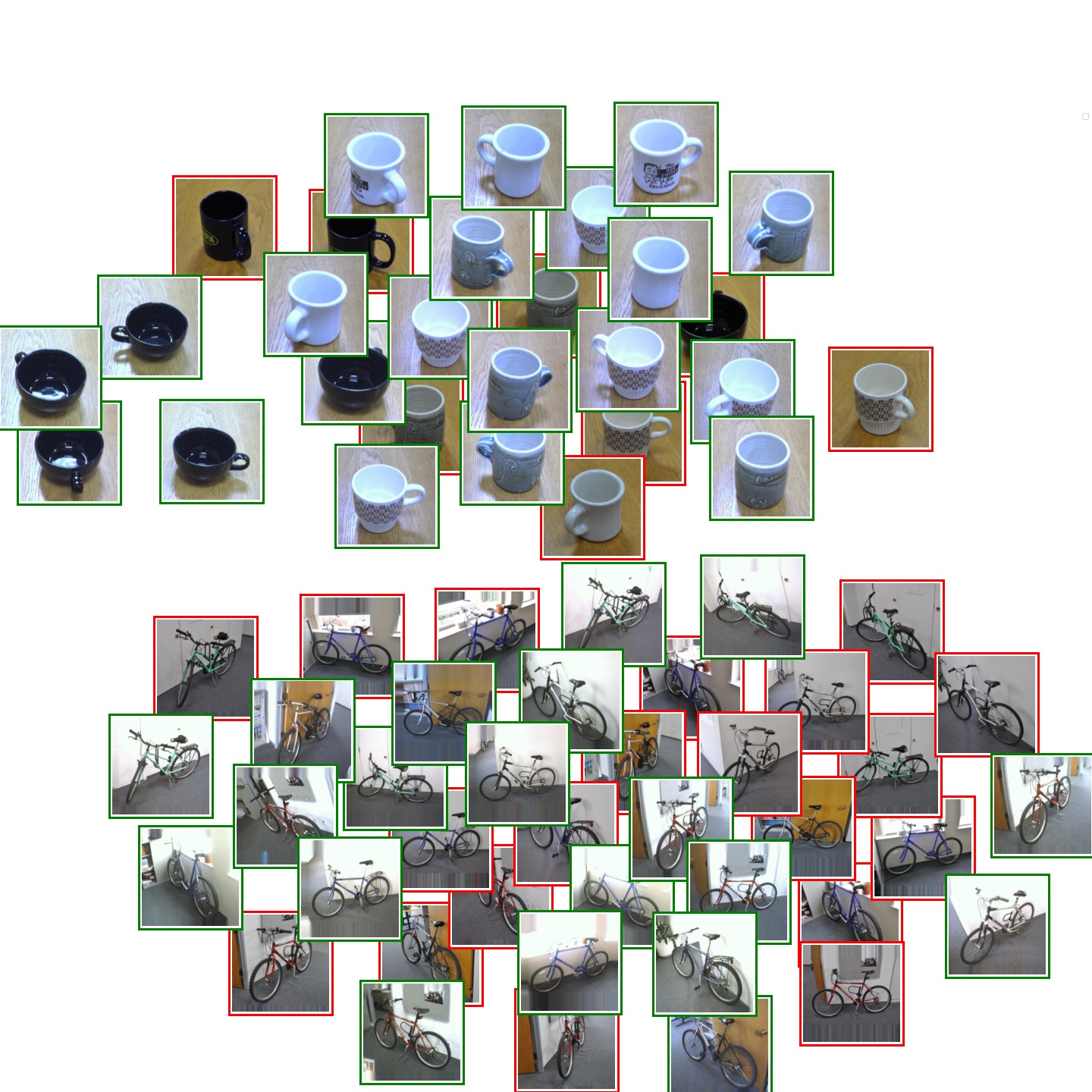}
        %\vspace{-4mm}
        \caption*{\small (d) Feature space from SGF with $t=0.6$.}
        \end{minipage}
        %\vspace{-6mm}
    \end{center}
    \end{experiment}
    }

    \FloatBox*[\linewidth]{
    \vspace{-5mm}
    \begin{experiment}[Shallow vs. Deep Features] \label{ex:tl:table} \normalfont
	\Cref{ex:tl:vgg} demonstrates the promising performance in visual
	domain adaption by enhancing (Grassmann) geodesic-flow methods with deep feature
	extraction. Inspired by the result, a more comprehensive investigation is
	carried out in this experiment based on $10$ image categories (instead of $2$ in
	\Cref{ex:tl:vgg}) from multiple domains A: \texttt{Amazon}, W: \texttt{WebCam}, D: \texttt{DSLR}
	and C: \texttt{Caltech-256}. Their different combinations create multiple transfer scenarios
	e.g., A $\to$ C means A is the source and $C$ is the target domains
	We compare shallow and deep feature extraction using the
	SURF and the VGG-16 model pretrained on the ImageNet. We trained
	four support vector machine (SVM) classifiers using the SURF or VGG16 features
	in the source domain with or without the help of GFK. Their different
	combinations correspond to the first four rows in
	\Cref{table:tl}.  We also include a model using VGG16 pretrain on the ImageNet
	and fine tuned in th source domain as both feature
	extractor and classifier in the fifth row. We report the
	average classification accuracies tested in the target. The results corroborate the two conclusions in
	\Cref{ex:tl:vgg}. First, compared with the direct-transfer approach, the
	application of GFK is effective in adapting to domain shifts by avoiding
	overfitting to the source domain. Second, geodesic-flow based transfer learning
	based on deep features substantially outperforms all other approaches with
	shallow features (SURF) or direct transfer. The performance gain is largest when
	the domain shift is large e.g., D $\to$ C. % Image classification relies on the
    \begin{center}
        \small
        \begin{tabularx}{\textwidth}{Xcccccc}
    \toprule[1.5pt]
        & A $\to$ C          & D $\to$ A          & D $\to$ C          & W $\to$ A          & W $\to$ C          & C $\to$ W          \\ \midrule[1.5pt]
           SURF+SVM & $ 36.20 \pm 2.83 $ & $ 30.15 \pm 3.34 $ & $ 29.18 \pm 3.01 $ & $ 29.55 \pm 4.38 $ & $ 27.78 \pm 3.91 $ & $ 22.69 \pm 3.67 $ \\ \midrule
     SURF+GFK+SVM
     %(PCA, SVM)
     & $ 33.25 \pm 2.91 $ & $ 29.25 \pm 5.18 $ & $ 27.50 \pm 2.92 $ & $ 24.15 \pm 5.13 $ & $ 23.68 \pm 3.78 $ & $ 25.50 \pm 4.65 $ \\ \midrule
          VGG+SVM & $ 79.38 \pm 2.74 $ & $ 75.08 \pm 6.15 $ & $ 63.55 \pm 4.55 $ & $ 69.55 \pm 5.82 $ & $ 59.40 \pm 4.99 $ & $ 78.00 \pm 5.37 $ \\ \midrule
    VGG+GFK+SVM
    %(PCA, SVM)
    & $ 84.65 \pm 2.28 $ & $ 81.68 \pm 3.75 $ & $ 77.72 \pm 3.82 $ & $ 76.65 \pm 3.89 $ & $ 72.70 \pm 3.90 $ & $ 77.62 \pm 5.23 $ \\ 
    \midrule
         VGG & $ 82.66 $          & $ 46.87 $          & $ 48.99 $          & $ 58.66 $          & $ 55.30 $          & $ 73.90 $          \\ 
\midrule
    \bottomrule[1.5pt]
        \end{tabularx}
        %\vspace{-4mm}
        \captionof{table}{Unsupervised domain adaptation with shallow and deep features. }
        \label{table:tl}
    \end{center}
    \vspace{-5mm}
    \end{experiment}
    }

    % Due to the high expressivity and representational power of neural nets, it is thus natural to ask if GFK is still necessary under this case.
    % With feature extracted from a pre-trained model, these methods can be even more effective.

\subsection{Deep Neural Networks on the Grassmann Manifold}
\label{sec:deep:net}

\begin{figure*}
        \centering
        \def\svgwidth{0.9\textwidth}
        \input{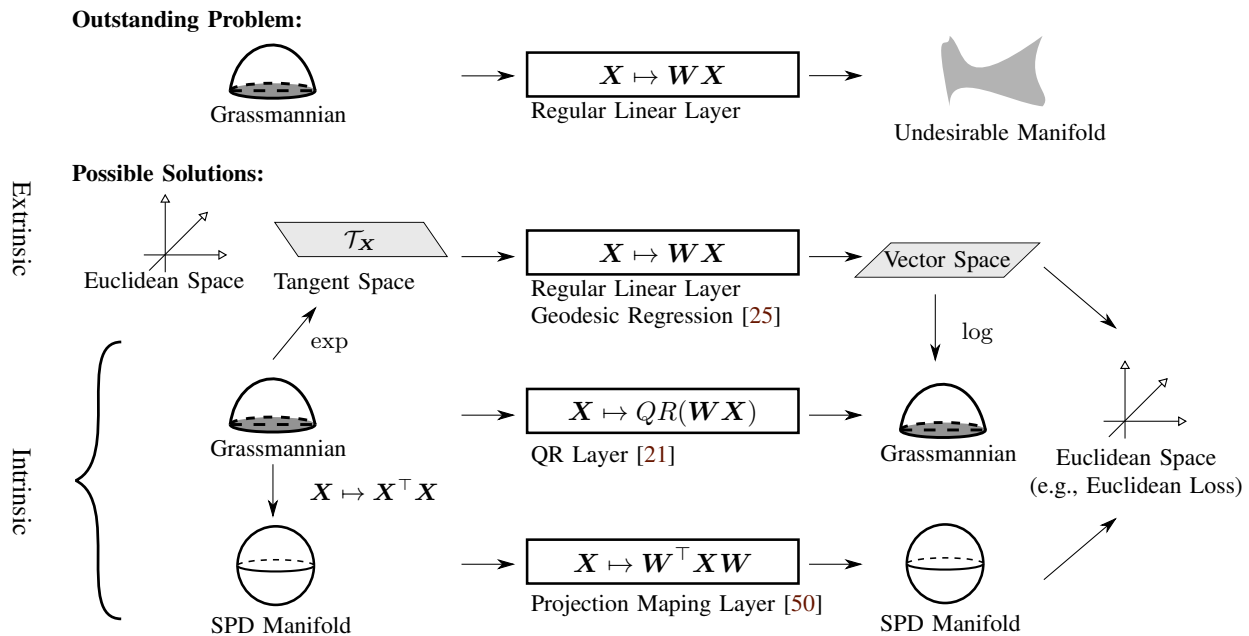}
        \caption{The problem of existing neural networks and different approaches for constructing mapping network layers on the Grassmann manifold. }
         \label{fig:grnet_flow}
    \end{figure*}

    In the preceding section, the simple cascading of deep neural networks for
    feature extraction and Grassmannian learning for domain adaption shows promising
    results. This suggests the  direction of constructing  deep neural networks
    that directly operate on the Grassmann manifold, targeting applications with
    Grassmannian input data (e..g, image sets \cite{Huang:2016:GRNET}) or subspaces
    as output \cite{Lohit:2017:LIR}. Recently, some progress has been made in this
    direction. In particular, several building blocks for constructing deep neural
    networks on manifolds (Grassmann or general Riemann manifolds) have been
    developed \cite{Harandi:2016:STBP, Masci:2015:GC}.  The framework of
    matrix back-propagation was first proposed in \cite{Ionescu:2015:MBP}, which
    generalizes the conventional vector version for gradient calculation. Advanced
    matrix analysis tools including partial derivatives of decompositions are
    deployed in developing the framework. Subsequently, the framework was further
    developed in \cite{Harandi:2016:STBP} for tackling matrices under orthogonality
    constraints and thus lying on the Stiefel manifolds. In this line of research,
    operations in the conventional vector-based neural networks such as pooling and
    normalization layers are redesigned for handling manifold-type data and signals.
    For instance, inspired by the success of convolutional layers, the notion of
    geodesic convolution was proposed in \cite{Masci:2015:GC}. Its strength lies in
    tasks such as establishing shape correspondence and shape retrieval in
    pose-recognition. Nevertheless, neural network layers thus designed cannot be
    directly extended to support Grassmann deep learning as most operations (either
    linear or nonlinear transformations) are incapable of preserving the geometry of
    the Grassmann manifolds. To enforce the Grassmannian constraints on the
    component network layers, a number of methods have been proposed which fall into
    one of two major approaches, namely the \emph{intrinsic} and \emph{extrinsic}
    approaches, depending on whether a method requires embedding a manifold in a
    higher-dimensional Euclidean space as elaborated in the sequel. We illustrate
    the principles and key operations for these approaches in Fig.
    \ref{fig:grnet_flow}. They are discussed in the following subsections. Before
    that, we remark that despite some initial progress, the field of Grassmannian
    deep learning is nascent field and potentially a gold mine of research
    opportunities.

    \subsubsection{Intrinsic Grassmann Deep Learning} There exists two methods
    for constructing deep neural networks that operate {\bf intrinsically} on
    the Grassmann manifold. Both methods are depicted in the lower half of Fig.
    \ref{fig:grnet_flow}. The first is to identify points on the Grassmann
    manifold by projecting it to the corresponding \emph{symmetric positive
    definite} (SPD) manifold via the mapping $\X \mapsto \tp{\X}\X$.  As a
    result, the linear transformation between layers in a neural network is
    modified from $\X \mapsto \tp{\W}\X$ to $\tp{\W}\X\W$ for some learnable
    weights $\W$ in the projection layer as proposed in \cite{Huang:2017:SPD}.
    Note that the output $\tp{\W}\X\W$ is on the SPD manifold whenever $\X$ is.
    Thereby, the structure of the SPD manifold is preserved.

\begin{figure*}
        \centering
        \def\svgwidth{0.8\textwidth}
        \input{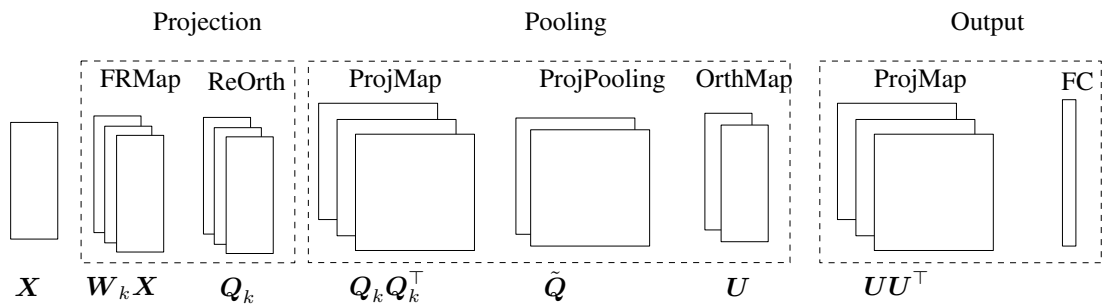}
	\caption{One design of neural network architecture on Grassmann
	    manifolds (reproduced from \cite{Huang:2016:GRNET}). $\X$ denotes
	    some Grassmannian data sample fed into the network; $\W_{k}$ is the
	    weights for a particular filter of full rank transform; $\Q_k$ is
	    the $\Q$ component of the QR decomposition on $\W_{k}\X$;
	    $\tilde{\Q}$ is the result
        after average pooling on SPD, i.e., the arithmetic average of $\Q_k \tp{\Q_k}$ for some filters; $\U$
        is the eigenspace of $\tilde{\Q}$ extracted by eigen decomposition; finally, the output layer is the vectorized
        output of the final projection layer.}
         \label{fig:grnet}
    \end{figure*}

    The other method based on the intrinsic approach is to design specific
    projection layers that preserve the manifold geometry. The so called ``GrNet"
    (Grassmann network) proposed by \cite{Huang:2016:GRNET} utilizes the QR
    decomposition to devise layers that output the $\Q$ component of the result
    after some linear transformations, which in effect restricts the output to lie
    on some Grassmann manifold. Then conventional Euclidean loss functions can be
    concatenated after a projection layer and used for gradient-descent based
    training. The key mathematical tools in the design include the differentiation
    of QR and eigen decompositions for back-propagation whose details can be found
    in \cite{Huang:2016:GRNET,Ionescu:2015:MBP}.  The architecture of the GrNet is
    depicted in Fig. \ref{fig:grnet}.

    The initial attempts on developing customized deep neural network for Grassmannian data
    such as GrNet \cite{Huang:2016:GRNET} or SPDNet \cite{Huang:2017:SPD} have
    yielded promising performance in tasks such as video-based classification,
    emotion classification and activity recognition. Further investigations in
    this direction are necessary to fully leverage the rich literature of signal
    processing on Grassmann manifolds and reduce the generalization error by
    improving the network architecture.

    \subsubsection{Extrinsic Grassmann Deep Learning}
	The extrinsic approach preserves the Grassmannian geometry by projecting
	Grassmannian data onto the tangent space at some chosen origin as depicted in
	the upper half of Fig. \ref{fig:grnet_flow}. Since the tangent space is a vector
	space, one may deploy a neural network with regular linear layers to operate on
	tangent vectors. Specifically, a conventional deep neural network can be trained
	to learn a mapping from input data space (typically Euclidean space) to the
	tangent space of some Grassmann manifold. The result is then projected back on
	the manifold by logarithmic map (see Section \ref{sec:pre:geo}). This approach
	also enables us to generate Grassmannian outputs from Euclidean data, which is
	referred to as the \emph{geodesic regression} in \cite{Lohit:2017:LIR}. Methods
	based on the extrinsic approach have been demonstrated to be effective in
	learning a subspace-invariant representation of illumination spaces in images.

\section{Applications}
\label{sec:applications}
    In this section, several canonical applications of Grassmannian learning are discussed,
    including image-set/video based classification, wireless communications, and
    recommender systems.

    \paragraph{Image-set/Video Based Recognition and Classification}

\begin{figure*}
    \centering
    \begin{subfigure}[b]{0.5\linewidth}
        \centering
        \includegraphics[width=\textwidth]{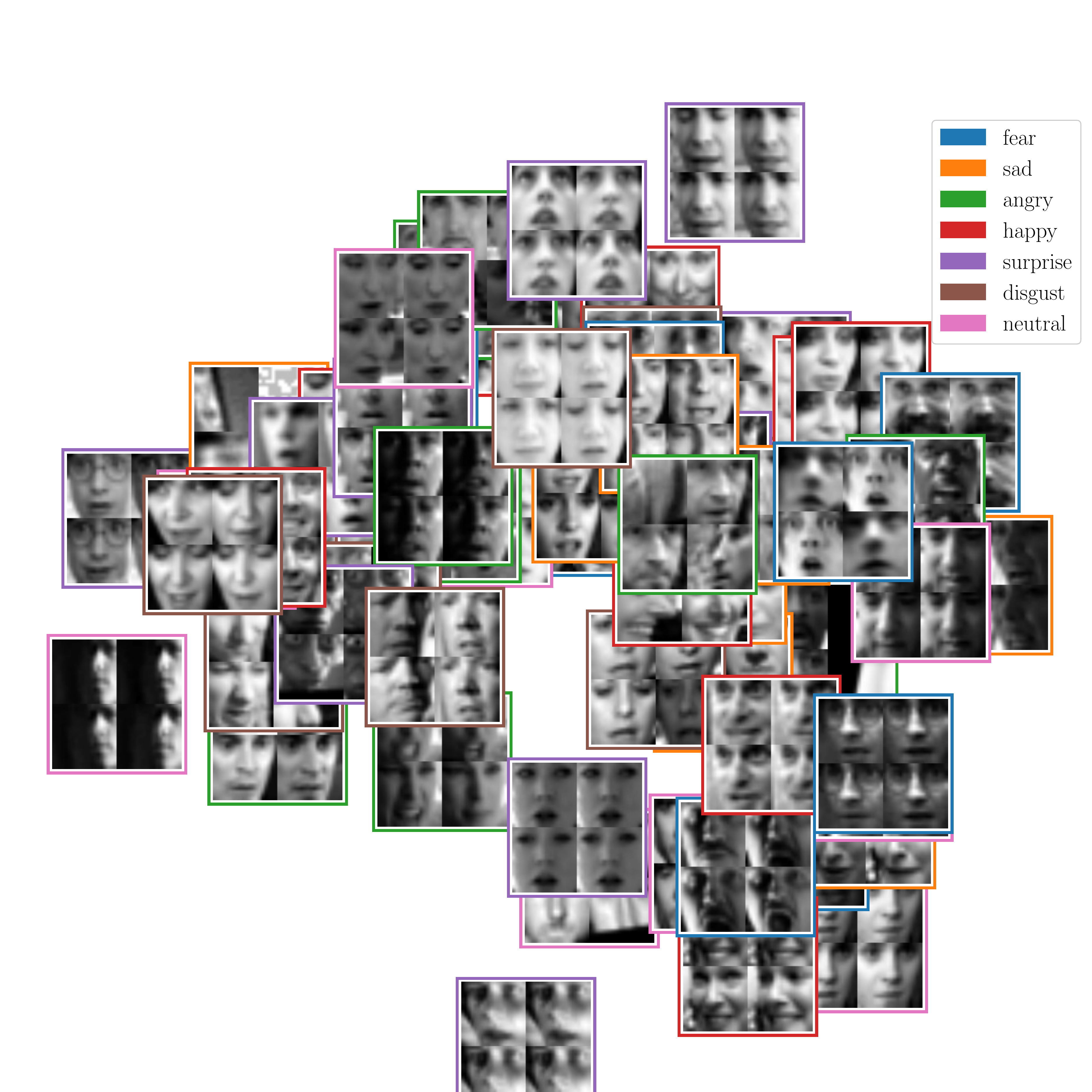}
        \caption{\emph{Raw Data.}}
        \label{fig:gda:original}
    \end{subfigure}%
    \begin{subfigure}[b]{0.5\linewidth}
        \centering
        \includegraphics[width=\textwidth]{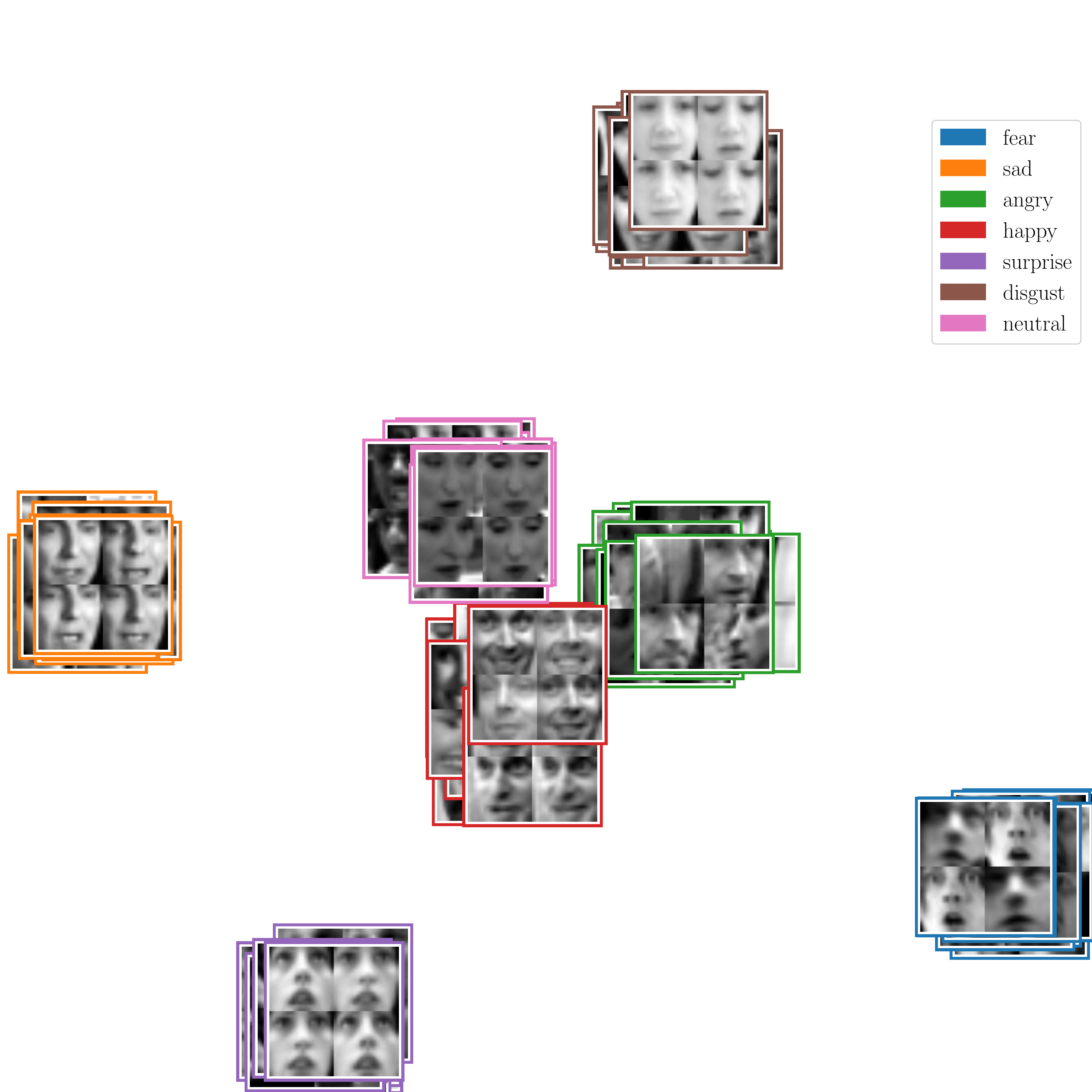}
        \caption{\emph{Dimensionality reduction via GDA.}}
        \label{fig:gda:gda}
    \end{subfigure}
    \caption{Emotion classification using Grassmannian learning (GDA) from the AFEW Dataset (best viewed in colour). }
    \label{fig:gda}
\end{figure*}

    Image/video based recognition or classification is a classical
    problem in computer vision. Prior to the resurgence of deep neural networks,
    such tasks usually rely on handcrafted feature extractors such as SIFT, HoG
    or SURF and a variety of dimensionality reduction techniques. Learning in
    computer vision, by nature, is  closely related to linear subspaces (or Grassmann manifolds). For instance,
    in a properly chosen subspace, the features of a subject can be invariant
    under different poses or illuminations and differentiable from those of
    another subject. Then an image/video recognition problem can be formulated
    as a discriminant learning problem on the Grassmann manifold
    (\Cref{sec:trad:disc}). As a concrete example, we consider image-set based
    emotion classification problem, where we use the dataset, \emph{acted facial
    expressions in the wild} (AFEW) \cite{AFEW:1,AFEW:2}, to demonstrate the
    algorithm. This dataset contains video clips categorized under sever
    different emotions (happy, sad, angry, fear, neutral, exciting, surprised).
    In Fig. \ref{fig:gda}, the results obtained from GDA introduced in
    Section~\ref{sec:trad:disc} are presented. The well-known $t$-SNE
    \cite{Maaten:2008:TSNE} algorithm to applied to visualize the raw data in
    Fig. \ref{fig:gda:original} and the discriminant representation learnt from
    GDA in \cref{fig:gda:gda}. One can observe that GDA recovers the
    discriminative representations of emotions hidden in high-dimensional raw
    data.

    \paragraph{Intelligent Multiple-Input Multiple-Output (MIMO) Communication}

    \begin{figure*}
        \centering
        %\small
        \resizebox{\textwidth}{!}{%
        \def\svgwidth{1.6\linewidth}
        \input{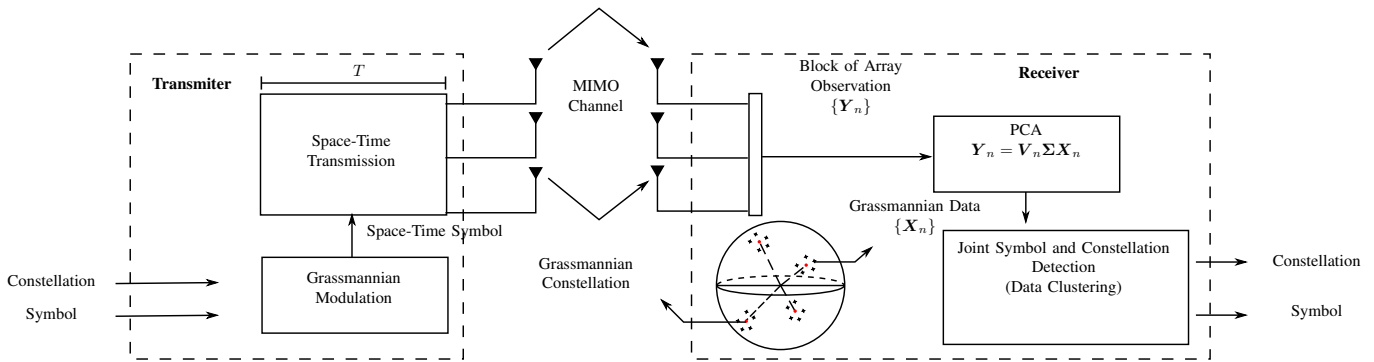}}
        \caption{Automatic recognition of space-time constellation by learning on the Grassmann manifold for in an intelligent MIMO communication system.}
         \label{fig:grassmimo}
    \end{figure*}

    MIMO communications has been a key physical technology driving the evolution of
    wireless systems from 1G to 5G. Its feature is to leverage spatial degrees of
    freedom generated by antenna arrays to scale up data rates by spatial
    multiplexing or improve link reliability by spatial diversity
    \cite{Gesbert:2010:MIMO}. Grassmann manifolds have been exploited in different
    areas of MIMO communication, most notable applications are  non-coherent (Grassmann) MIMO modulation \cite{Hochwald:2000:Unitary} and quantization of precoder feedback
    \cite{love2008overview}.
For  precoder  quantization,  precoder codebooks  can be generated by quantizing a Grassmann manifold, the space of unitary  precoder matrices,  using e.g., the Grassmann K-means algorithm.
     In non-coherent MIMO modulation, a Grassmann constellation consists of a set
    of points on a Grassmann manifold that are computed e.g., from subspace packing
    \cite{Zheng:2002:GRASS}.  In practice, non-coherent MIMO had not been as popular  as coherent MIMO as the former cannot scale the data rate by spatial
    multiplexing as the latter. Nevertheless, recent years have seen the resurgence
    of non-coherent MIMO in research on next-generation low-latency low-rate
    machine-type communication as the technique requires no channel training and is
    robust against fading \cite{panigrahi2017feasibility}. Recently, targeting
    next-generation intelligent MIMO receiver, a framework of automatic recognition
    of space-time constellation is developed in \cite{GrassMIMO} leveraging
    algorithms for unsupervised learning on Grassmann manifold originally developed
    for computer vision.  The system proposed in \cite{GrassMIMO} is illustrated in
    Fig. \ref{fig:grassmimo}.

\begin{figure}[tp]
        \centering
        \includegraphics[width=0.35\textwidth]{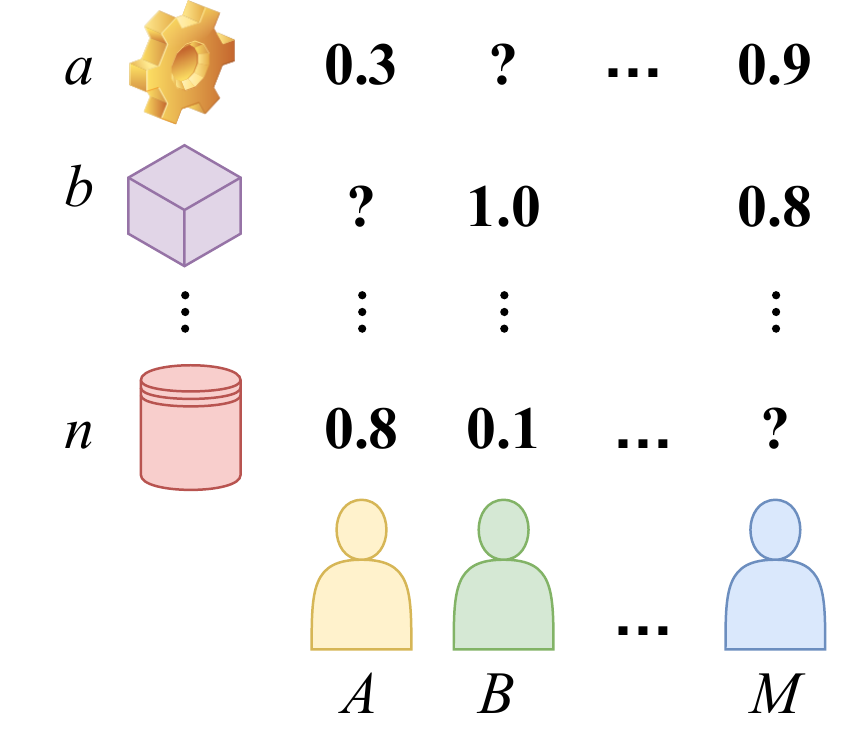}
        \caption{The matrix-completion problem in recommender systems.
        Each column of the matrix corresponds to a user, and each row to an item. The entries specify
    the preference of a user on an item. Under a low-rank constraint on the matrix, the recommender system can predict a  missing entry to provide recommendation to a new user. }
         \label{fig:recsys}
    \end{figure}

\paragraph{Recommender Systems}
    One practical use case of the low-rank representation learning is to build
    recommender systems using the Grassmannian optimization method for  low-rank matrix completion as  introduced in
    \Cref{sec:trad:lrr} \cite{Candes:2009:MCCV}. The preferences of users on
    the items can be formed by a \emph{preference matrix}, where the rows and
    columns represent items and users. The entries are the scores of preferences and
    the missing entries correspond to unavailable data. For example, in the NetFlix
    Challenge \cite{Bennett:2007:NETFLIX}, the preferences are the motive ratings by
    viewers. Then the missing entries can be
    reconstructed using techniques for matrix completion. An illustration of a
    recommender system is given in Fig. \ref{fig:recsys}.

\section{Concluding Remarks}
\label{sec:conclusion}

    In this paper, we introduced the preliminaries and the shallow and
    deep paradigms of Grassmannian learning. Relevant techniques have been
    demonstrated using a set of examples and applications covering areas including
    computer vision, wireless communications, and recommender systems. Despite the
    separation into different topics in this paper for the purpose of exposition,
    shallow and deep Grassmann learning cannot be treated as two separate
    areas. In contrast, they are interwound where techniques in the former area
    playing the role of building blocks of the latter. The latest paradigm,
    Grassmann deep learning, is a nascent but fast growing area with many
    research opportunities. Some of them are summarized as follows.

\begin{itemize}

    \item \emph{Embedding Geometry in Deep Neural Networks:}
	Recent years have seen growing interest in geometry-based deep neural networks,
with the potential to become a mainstream approach for improving accuracy
	and robustness of deep learning.  In this direction, Grassmannian learning
	techniques may play a key role as subspace structural
	information is embedded in data especially image sets or video. Despite some
	initial progress \cite{Huang:2016:GRNET,Herath:2017:ILS}, geometry based deep
	neural networks represents a paradigm shift where there are still many open
	challenges such as overcoming over-fitting by regularization on the Grassmann
	manifolds, devising more efficient optimizers for non-Euclidean layers, and
	leverage its power on acquiring intelligence from  real-world datasets.

    \item \emph{General Geometric Deep Learning:}
	Grassmannian deep learning, a  theme of this paper, lies in the general area of
	geometric deep learning. The area represents a new trend in the deep learning
	community, which involves deep learning from geometric data including not only
	images/videos but also other types of data such as 3-dimensional objects,
	graphic mashes, or social networks \cite{Bronstein:2017:GEODL}. The underpinning
	basic mathematical toolset is optimization on Riemann manifolds. This area is
	still largely uncharted.

    \item \emph{Robust Machine Learning:} One critical weakness of learning models, especially more flexible models
	such as deep neural networks, is their susceptibility to malicious adversarial perturbations
	that mislead the models to make incorrect decisions
	\cite{Szegedy:2013:ADV,GF:2014:ADV}. Grassmannian
	learning exhibits a certain degree of robustness against small perturbations
	\cite{Chikuse:2012:STAT}. The intuitive reason is that it is not easy for
	small perturbations to change one subspace to another. Hence it warrants
	further study on how to leverage Grassmann manifolds (or other Riemann
	manifolds) to devise more robust machine learning models.

\end{itemize}

    The fast growth of Grassmannian deep learning is assisted by the availability of
    high-performance software for Riemannian optimization, such as \texttt{ManOpt}
    \cite{MANOPT} for \texttt{MATLAB} and \texttt{pyManOpt}
    \cite{PYMANOPT} for \texttt{Python}. The software packages render implementation
    and testing of Grassmannian learning algorithms more accessible to general
    practitioners in signal processing.

    The paper is ended with a hope that this work provides an accessiable and
    inspiring introduction to the area of Grassmann machine learning. Besides an
    interesting read, readers will be equipped with adequate fundamentals to apply
    Grassmannian learning to novel scenarios and applications.

\footnotesize{
\bibliographystyle{IEEEtranN}
% Generated by IEEEtranN.bst, version: 1.14 (2015/08/26)

}

\clearpage

\end{document}